\documentclass[letterpaper]{article}
\usepackage{uai2019}
\usepackage[margin=1in]{geometry}
\usepackage{times}
\usepackage[T1]{fontenc}
\usepackage[latin1]{inputenc}
\usepackage[english]{babel}
\usepackage{amsopn}
\usepackage{url}
\usepackage{smallref}
\usepackage{algorithm}
\usepackage{algpseudocode}
\usepackage{comment}

\algrenewcommand\algorithmicindent{1.0em}

\usepackage[normalem]{ulem}

\usepackage{cite}		%

\usepackage{amsmath, amsfonts}
\usepackage{amssymb}
\usepackage{amsthm}
\usepackage[all]{xy}
\usepackage{cancel}
\usepackage{dsfont}
\usepackage{enumitem}
\usepackage{wasysym}
\usepackage{tikz}
\usepackage{stackrel}

\usetikzlibrary{arrows}
\usetikzlibrary{arrows.meta, calc}
\usetikzlibrary{shapes,positioning,decorations.markings}
\usetikzlibrary{decorations.pathreplacing}

\theoremstyle{plain}

\newtheorem{sa}{Theorem}[section]
\newtheorem{Thm}[sa]{Theorem}
\newtheorem{Lem}[sa]{Lemma}
\newtheorem{Prp}[sa]{Proposition}
\newtheorem{Cor}[sa]{Corollary}

\newtheorem{Def}[sa]{Definition}

\newtheorem{Not}[sa]{Notation}

\newtheorem{Rem}[sa]{Remark}

\newtheorem{Eg}[sa]{Example}

\newcommand{\id}{\mathrm{id}}

\newcommand{\CDcal}{\mathcal{CD}}
\newcommand{\Xcal}{\mathcal{X}}

\newcommand{\Scal}{\mathcal{S}}
\newcommand{\Lcal}{\mathcal{L}}
\newcommand{\Ical}{\mathcal{I}}

\newcommand{\obs}{\diameter} %

\newcommand{\sm}{\setminus}						
\newcommand{\ins}{\subseteq}

\newcommand*{\tuh}[1][]{\mathrel{\tikz [baseline=-0.25ex,-latex, #1] \draw [#1] (0pt,0.5ex) -- (1.3em,0.5ex);}}
\newcommand*{\hut}[1][]{\mathrel{\tikz [baseline=-0.25ex,latex-, #1] \draw [#1] (0pt,0.5ex) -- (1.3em,0.5ex);}}
\newcommand*{\huh}[1][]{\mathrel{\tikz [baseline=-0.25ex,latex-latex, #1] \draw [#1] (0pt,0.5ex) -- (1.3em,0.5ex);}}

\newcommand*{\rars}{\begin{array}{c}\huh\\[-10pt]\tuh\end{array}}
\newcommand*{\lars}{\begin{array}{c}\huh\\[-10pt]\hut\end{array}}

\newcommand*{\dars}{\begin{array}{c}\huh\\[-10pt]\tuh\\[-10pt]\hut\end{array}}

\renewcommand{\Pr}{\mathbb{P}}

\newcommand{\Pa}{\mathrm{Pa}} 		
 		
\newcommand{\Ch}{\mathrm{Ch}} 		
\newcommand{\Anc}{\mathrm{Anc}} 		
 
\newcommand{\Desc}{\mathrm{Desc}} 	

\newcommand{\CDist}{\mathrm{Cd}} 
 
\newcommand{\Pred}{\mathrm{Pred}} 
\newcommand{\Sc}{\mathrm{Sc}}
\newcommand{\NonDesc}{\mathrm{NonDesc}}
\DeclareMathOperator*{\Indep}{{\,\perp\mkern-12mu\perp\,}}
\DeclareMathOperator*{\nIndep}{{\,\not\mkern-1mu\perp\mkern-12mu\perp\,}}
\DeclareMathOperator*{\given}{|}

\DeclareMathOperator{\doit}{do}

\newcommand{\acy}{\mathrm{acy}}

\newcommand{\ioSCM}{ioSCM}
\newcommand{\ioSCMs}{ioSCMs}

\newcommand{\hyte}[2]{\emph{#2}}

\title{Causal Calculus in the Presence of Cycles, Latent Confounders and Selection Bias}

\author{ {\bf Patrick Forr\'e}\\
Informatics Institute\\University of Amsterdam\\The Netherlands\\
\url{p.d.forre@uva.nl}\\
\And
{\bf Joris M.~Mooij}\\
Informatics Institute\\University of Amsterdam\\The Netherlands\\
\url{j.m.mooij@uva.nl}\\
}

\begin{document}

\maketitle

\begin{abstract}
We prove the main rules of \emph{causal calculus} (also called \emph{do-calculus}) for \emph{i/o structural causal models (\ioSCMs)}, a generalization of a recently proposed general class of non-/linear structural causal models that allow for cycles, latent confounders and arbitrary probability distributions. 
We also generalize \emph{adjustment criteria and formulas} from the acyclic setting to the general one (i.e.\ \ioSCMs). Such criteria then allow to estimate (conditional) causal effects from observational data that was (partially) gathered under selection bias and cycles. This generalizes the \emph{backdoor criterion}, the \emph{selection-backdoor criterion} and extensions of these to arbitrary \ioSCMs.
Together, our results thus enable causal reasoning in the presence of cycles, latent confounders and selection bias.
Finally, we extend the ID algorithm for the identification of causal effects to \ioSCMs.

\end{abstract}

\setlist{nosep}

\section{INTRODUCTION}
Statistical models are governed by the rules of probability (e.g.\ sum and product rule), which link joint distributions with the corresponding (conditional) marginal ones. \emph{Causal models} follow additonal rules, which relate the observational distributions with the interventional ones. In contrast to the rules of probability theory, which directly follow from their axioms, the rules of \emph{causal calculus} need to be proven, when based on the definition of \emph{structural causal models} (SCMs). As SCMs will among other things depend on the underlying graphical structure (e.g.\ with or without cycles or bidirected edges, etc.), the used function classes (e.g.\ linear or non-linear, etc.) and the allowed probability distributions (e.g.\ discrete, continuous, singular or mixtures, etc.) the respective endeavour is not immediate.

Such a framework of causal calculus contains rules about when one can 1.) insert/delete observations, 2.) exchange action/observation, 3.) insert/delete actions; and about when and how to recover from interventions and/or selection bias (backdoor and selection-backdoor criterion), etc. (see \cite{Pearl93as,Pearl93bs,Pearl95s,Tian2002,SP06,SP06s,HV06s,Pearl09,PearlPaz13,SWR10s,BTP14s,PTKM15s,CB17s, CTB18s}).
While these rules %
have been extensively studied for \emph{acyclic} causal models, e.g.\ (semi-)Markovian models, which are attached to directed acyclic graphs (DAGs) or acyclic directed mixed graphs (ADMGs) (see \cite{Pearl93as,Pearl93bs,Pearl95s,Tian2002,SP06,SP06s,HV06s,Pearl09,PearlPaz13,SWR10s,BTP14s,PTKM15s,CB17s, CTB18s}), the case of causal models with \emph{cycles} stayed in the dark. %

To deal with cycles and latent confounders at the same time in this paper we will introduce the class of 
\emph{input/output structural causal models (\ioSCMs)}, a ``conditional'' version of the recently proposed class of \emph{modular structural causal models (mSCMs)} (see \cite{FM17,FM18}) to also include ``input'' nodes that can play the role of parameter/context/action/intervention nodes.
\ioSCMs\  have several desirable properties: 
They allow for arbitrary probability distributions, non-/linear functional relations, latent confounders and cycles. They can also model non-/probabilistic external and probabilistic internal nodes in one framework.
The cycles are modelled in a least restrictive way such that the class of \ioSCMs\ still becomes closed under arbitrary marginalizations and interventions. 
All causal models that are based on acyclic graphs like DAGs, ADMGs or mDAGs (see \cite{Richardson03,Eva15}) %
can be interpreted as special acyclic \ioSCMs. Besides feedback over time \ioSCMs\ can also express instantaneous and equilibrated feedback under the made model assumptions (e.g.\ the ODEs in \cite{MJS13s,BM18ss}).
All models where the non-trivial cycles are ``contractive'' (negative feedback loops, see \cite{FM18}) are \ioSCMs\ without further assumptions. Thus \ioSCMs\ generalize all these classes of causal models in one framework, which goes beyond the acyclic setting and also allows for conditional versions of those (e.g.\ CADMGs), expressed via external non-/probabilistic ``input'' nodes. 
Also the \emph{generalized directed global Markov property} for mSCMs (see \cite{FM17,FM18}) generalizes to \ioSCMs, i.e.\ \ioSCMs\ entail the conditional independence relations that follow from
the \emph{$\sigma$-separation criterion} in the underlying graph, where \emph{$\sigma$-separation} generalizes the usual 
d-separation (also called m- or m$^*$-separation, see \cite{Pearl86b,VerPea90a,Pearl09,Eva15,Richardson03}) from acyclic graphs to directed mixed graphs (DMGs) (and even HEDGes \cite{FM17} and $\sigma$-CGs \cite{FM18}) with or without cycles in a non-naive way.%

This paper now aims at proving the mentioned main rules of causal calculus for \ioSCMs\ and derive adjustment criteria with corresponding adjustment formulas like generalized (selection-)backdoor adjustments. We also provide an extension of the ID algorithm for the identification of causal effects to the \ioSCM\ setting, which reduces to the usual one in the acyclic case.

The paper is structured as follows: We will first give the precise definition of \ioSCMs\ closely mirroring mSCMs from \cite{FM17,FM18}. We will then review $\sigma$-separation and generalize its criterion from mSCMs (see \cite{FM17,FM18}) to \ioSCMs. As a preparation for the causal calculus, which relates observational and interventional distributions, we will then show how one can extend a given \ioSCM\ to one that also incorporates additional interventional variables indicating the regime of interventions on the observed nodes. We will then show how the rules of causal calculus directly follow from applying the $\sigma$-separation criterion to such an extended \ioSCM. We then derive the mentioned general adjustment criteria with corresponding adjustment formulas. Finally, we introduce the right definitions for \ioSCMs\ to extend the ID algorithm for the identification of causal effects to the general setting.

\section{INPUT/OUTPUT STRUCTURAL CAUSAL MODELS}

In this section we will define \emph{input/output structural causal models} (\ioSCMs), which can be seen as a ``conditional'' version of modular structural causal models (mSCMs) defined in \cite{FM17,FM18}.
We will then construct marginalized \ioSCMs\ and intervened \ioSCMs. 
To allow for cycles we first need to introduce the notion of loop of a graph and its strongly connected components.

\begin{Def}[Loops]
Let $G=(V,E)$ be a directed graph (with or without cycles).
\begin{enumerate}
  \item A set of nodes $S \ins V$ is called a \hyte{loop}{loop} of $G$ if
	 for every two nodes $v_1,v_2 \in S$ there are two directed walks $v_1 \tuh \cdots \tuh v_2$ and  $v_2 \tuh \cdots \tuh v_1$ in $G$ such that  all the intermediate nodes are also in $S$ (if any). The sets $S=\{v\}$ are also considered as loops (independent of $v \tuh v \in E$ or not). 
	\item The set of loops of $G$ is written as $\Lcal(G)$.
\item The \hyte{Sc}{strongly connected component} of $v$ in $G$ is defined to be:
 $\quad \Sc^G(v):= \Anc^G(v) \cap \Desc^G(v).$ %
\item The set of strongly connected components is $\Scal(G)$.
\end{enumerate}
\end{Def}

\begin{Rem}
Let $G=(V,E)$ be a directed graph.
\begin{enumerate} 
\item We always have $v \in \Sc^G(v)$ and $\Sc^G(v) \in \Lcal(G)$.
\item If $G$ is acyclic then: $\Lcal(G)=\{\{v\} \,|\, v \in V\}$.
\end{enumerate}
\end{Rem}

In the following all spaces are meant to be equipped with $\sigma$-algebras and all maps to be measurable.
Whenever (regular) conditional distributions occur we implicitly assume standard measurable spaces (to ensure existence).

\begin{Def}[Input/Output Structural Causal Model]
\label{iSCM-def}
An \emph{input/output (i/o) structural causal model (\ioSCM)} by definition consists of:
\begin{enumerate}
\item a set of nodes $V^+=V \dot \cup U \dot \cup J$, where elements of $V$ correspond to output/observed variables, elements of $U$ to probabilistic latent variables and elements of $J$ to input/intervention variables.
\item an observation/latent/action space $\Xcal_v$ for every $v \in V^+$, $\Xcal:=\prod_{v \in V^+}\Xcal_v$,
\item a product probability measure $\Pr_U=\bigotimes_{u \in U} \Pr_u$ on the latent space $\Xcal_U:=\prod_{u \in U} \Xcal_u$,
\item a directed graph structure $G^+=(V^+, E^+)$
 with the properties: 
	\begin{enumerate}
	  \item $V = \Ch^{G^+}(U \cup J)$,   %
		\item $\Pa^{G^+}(U \cup J)=\emptyset$,
	\end{enumerate}
	where $\Ch^{G^+}$ and $\Pa^{G^+}$ stand for children and parents in $G^+$, resp.,\footnote{To have a ``reduced'' form of the latent space one can in addition impose the condition: $\Ch^{G^+}(u_1) \nsubseteq \Ch^{G^+}(u_2)$ for every two distinct $u_1,u_2 \in U$. This can always be achieved by gathering latent nodes together if $\Ch^{G^+}(u_1) \subseteq \Ch^{G^+}(u_2)$.}
\item a system of causal mechanisms $g=(g_S)_{\substack{S \in \Lcal(G^+)\\ S \ins V}}$: 
 \[ g_S: \; \prod_{v \in \Pa^{G^+}(S)\sm S}\Xcal_v \to \prod_{v \in S}\Xcal_v,%
\footnote{Note that the index set runs over all ``observable loops'' $S \ins V$, $S \in \Lcal(G^+)$, not just the sets $\{v\}$ for $v\in V$.}%
\]
that satisfy the following \emph{global compatibility} conditions: For every nested pair of loops $S' \ins S \ins V$ of $G^+$ and every element
$x_{\Pa^{G^+}(S) \cup S} \in \prod_{ v\in \Pa^{G^+}(S) \cup S} \Xcal_v$ we have the implication:
\[\begin{array}{rcrcl} 
 && g_S(x_{\Pa^{G^+}(S) \sm S}) &=& x_S \\
 &\implies& g_{S'}(x_{\Pa^{G^+}(S') \sm S'})&=&x_{S'},  \end{array}\]
where $x_{\Pa^{G^+}(S') \sm S'}$ and $x_{S'}$ denote the corresponding components of $x_{\Pa^{G^+}(S) \cup S}$.
\end{enumerate}
The \ioSCM\ will be denoted by $M=(G^+,\Xcal,\Pr_U,g)$.
\end{Def}

\begin{Def}[Modular structural causal model, see \cite{FM17,FM18}]
A \emph{modular structural causal model (mSCM)} is an \ioSCM\ without input nodes, i.e.\ $J=\emptyset$. %
\end{Def}

\begin{Rem}[Composition of \ioSCMs]
\label{composition-ioSCMs}
Consider two \ioSCMs\ $M_1$, $M_2$ and an identification of subsets $I_1 \ins V_1^+$ with $I_2 \ins J_2$ and maps 
$g_{i_2}:\, \Xcal_{i_1} \to \Xcal_{i_2}$, for $i_1$ corresponding to $i_2$, e.g.\ $g_{i_2}=\id$ if possible. We can now ``glue'' them together to get a new \ioSCM\ $M_3$ given by 
$V_3 := V_1 \dot\cup V_2 \dot\cup I_2$, $U_3 := U_1 \dot \cup U_2$, $J_3 = J_1 \dot\cup J_2 \sm I_2$ and $G^+_3 := G_1^+ \cup G_2^+$, where we add the
the edges $i_1 \tuh i_2$, and the mechanisms $g_{i_2}$ and $\Pr_{U_3}:=\Pr_{U_1} \otimes \Pr_{U_2}$.
\end{Rem}

\begin{Eg}[Constructing mSCMs from \ioSCMs]
Given an \ioSCM\ $M=(G^+,\Xcal,\Pr_U,g)$ with graph $G^+=(V \dot \cup U\dot \cup J, E^+)$ we can construct a well-defined mSCM by specifying a product distribution $\Pr_J:=\bigotimes_{j \in J} \Pr_j$ on $\Xcal_J:=\prod_{j \in J} \Xcal_j$ and following \ref{composition-ioSCMs}
with $M_1$ with only $U_1:=J_2$ without any edges and gluing maps $g_i:=\id$.
\end{Eg}

The actual joint distributions on the observed space $\Xcal_V$ and thus the random variables %
attached to any \ioSCM\ will be defined in the following.

\begin{Def}
\label{iSCM-variables}
Let $M=(G^+,\Xcal,\Pr_U,g)$ be an \ioSCM\ with $G^+=(V \dot \cup U\dot \cup J, E^+)$. 
The following constructions will depend on the choice  of a fixed value $x_J \in \Xcal_J$.
\begin{enumerate}
\item The latent variables are given by $(X_u)_{u \in U} \sim \Pr_U$, i.e.\ by the canonical projections $X_u :\, \Xcal_U \to \Xcal_u$, which are jointly $\Pr_U$-independent. We put $X_u^{\doit(x_J)}:=X_u$, i.e., independent of $x_J$.
\item For $j\in J$ we put $X_j^{\doit(x_J)}:=x_j$, the constant variable given by the $j$-component of $x_J$.
	\item The observed variables $(X_v^{\doit(x_J)})_{v \in V}$ are inductively defined by:
	 \[X_v^{\doit(x_J)} := g_{S,v}\big((X_w^{\doit(x_J)})_{w \in \Pa^{G^+}(S)\sm S}\big),\] 
	where $S:=\Sc^{G^+}(v):=\Anc^{G^+}(v) \cap \Desc^{G^+}(v)$ and 
	where the second index $v$ refers to the $v$-component of $g_S$. 
	The induction is taken over any topological order of the strongly connected components of $G^+$, which always exists (see \cite{FM17}).
		\item By the compatibility condition for $g$ we then have that for every $S \in \Lcal(G^+)$ with $S \ins V$ the following equality holds:
	\[X_S^{\doit(x_J)} = g_{S}(X_{\Pa^{G^+}(S)\sm S}^{\doit(x_J)}), \]
	where we put $\Xcal_A:=\prod_{v \in A} \Xcal_v$ and $X_A:=(X_v)_{v \in A}$ for subsets $A$.
	\item We define the family of conditional distributions:
	\begin{eqnarray*}
	&& \Pr_U(X_A | X_B, X_J=x_J) \\
	&:=&\Pr_U(X_A | X_B, \doit(X_J=x_J)) \\
	&:=& \Pr_U(X_A^{\doit(x_J)} | X_B^{\doit(x_J)}),
	\end{eqnarray*}
	for $A,B \ins V$ and $x_J \in \Xcal_J$. Note that in the following we will use the $\doit$ and the $\doit$-free notation (only) for the $J$-variables interchangeably.
	\item If we, furthermore, specify a product distribution $\Pr_J=\bigotimes_{j \in J} \Pr_j$ on $\Xcal_J$, then we get a joint distribution
	 $\Pr$ on $\Xcal_{V \cup J}$ by setting:
	$$\Pr( X_V,X_J):= \Pr_U(X_V|\doit(X_J)) \otimes \Pr_J(X_J).$$ 
\end{enumerate}	
\end{Def}
\vspace{0.2cm}

\begin{Rem}
\label{mSCM-all-subsets}
Let $M=(G^+,\Xcal,\Pr_U,g)$ be an \ioSCM\ with $G^+=(V \dot \cup U \dot \cup J, E^+)$.
For every subset $A \ins V$ we get a well-defined map
	 $\; g_A:\, \Xcal_{\Pa^{G^+}(A)\sm A} \to \Xcal_A,$
	by recursively plugging in the $g_S$ into each other for the biggest occuring loops $S \ins A$ by the same arguments as before.
These then are all globally compatible by construction and satisfy:
	\[X_A^{\doit(x_J)} = g_{A}(X_{\Pa^{G^+}(A)\sm A}^{\doit(x_J)}). \]
\end{Rem}

Similar to mSCMs (see \cite{FM17,FM18}) we can define the marginalization of an \ioSCM.

\begin{Def}[Marginalization of \ioSCMs]
\label{mSCM-marg}
Let $M=(G^+,\Xcal,\Pr,g)$ be an \ioSCM\ with $G^+=(V \dot \cup U \dot \cup J, E^+)$ and $W \ins V$ a subset.
The \emph{marginalized} \ioSCM\ $M^{\sm W}$ w.r.t.\ $W$ can be defined by plugging the functions $g_S$ related to $W$ into each other.
For example, when marginalizing out $W=\{w\}$ we can define (for the non-trivial case $w \in  \Pa^{G^+}(S) \sm S$):
	  \[\begin{array}{l} g_{S',v}(x_{\Pa^{(G^+)^{\sm W}}(S') \sm S'}) :=\\
		g_{S,v}\big(x_{\Pa^{G^+}(S) \sm (S \cup \{w\})}, g_{\{w\}}(x_{\Pa^{G^+}(w)\sm\{w\} })\big),  
\end{array}\]
  where $(G^+)^{\sm W}$ is the marginalized graph of $G^+$ (see Supplementary Material \ref{dmg}), $S' \ins V^{\sm W}:=V\sm W$ is any loop of $(G^+)^{\sm W}$ and $S$ the corresponding induced loop in $G^+$. 
\end{Def}

Similar to mSCMs (see \cite{FM17,FM18}) we now define what it means to intervene on \emph{observed} nodes in an \ioSCM.

\begin{Def}[Perfect interventions on \ioSCMs]
\label{mSCM-intv}
Let $M=(G^+,\Xcal,\Pr,g)$ be an \ioSCM\ with $G^+=(V \dot \cup U \dot \cup J, E^+)$. Let $W \ins V \cup J$ be a subset.
 We  then define the \emph{post-interventional} \ioSCM\ $M_{\doit(W)}$ w.r.t.\ $W$: %
\begin{enumerate}
\item Define the graph $G^+_{\doit(W)}$ by removing all the edges $v \tuh w$ for all nodes $w \in W$ and $v \in \Pa^{G^+}(w)$.
\item Put $V_{\doit(W)}:=V \sm W$ and $J_{\doit(W)}:= J \cup W$.
\item Remove the functions $g_S$ for loops $S$ with $S \cap W \neq \emptyset$. %
\end{enumerate}
The remaining functions then are clearly globally compatible and we get a well-defined \ioSCM\ $M_{\doit(W)}$.

\end{Def}
\section{CONDITIONAL INDEPENDENCE}

Here we generalize conditional independence for structured families of distributions. The main application will be the distributions 
$\left(\Pr_U(X_V|\doit(X_J=x_J)) \right)_{x_J \in \Xcal_J}$ coming from \ioSCMs, but the following definition might be of more general importance. 

\begin{Def}[Conditional independence]
\label{def-cond-ind}
Let $\Xcal_V:=\prod_{v \in V} \Xcal_v$ and $\Xcal_J:=\prod_{j \in J} \Xcal_j$ be product spaces and
$$ \Pr:=\left( \Pr_V(X_V|x_J) \right)_{x_J \in \Xcal_J}$$ 
a family of distributions on $\Xcal_V$ (measurably\footnote{We require that for every measurable $F \ins \Xcal_V$ the map $\Xcal_J \to [0,1]$ given by $x_J \mapsto \Pr_V(X_V \in F|x_J)$ is measurable. Such families of distributions are also called \emph{channels} or \emph{(stochastic) Markov (transition) kernels} (see \cite{Kle14}).}) parametrized by $\Xcal_J$.
For subsets $A,B,C \ins V \dot \cup J$ we write:
$$ X_A \Indep_{\Pr} X_B \given X_C$$
if and only if for \emph{every} product distribution $\Pr_J=\bigotimes_{j \in J} \Pr_j$ on $\Xcal_J$ we have:
$\quad X_A \Indep_{\Pr_{V\cup J}} X_B \given X_C,\quad \text{ i.e.: }$
$$ \Pr_{V \cup J}(X_A|X_B,X_C) = \Pr_{V \cup J}(X_A|X_C) \quad \Pr_{V \cup J}\text{-a.s.,} $$
where $\Pr_{V \cup J}(X_{V\cup J}):=\Pr_V(X_V|X_J)\otimes\Pr_J(X_J)$ is the distribution given by $X_J \sim \Pr_J$ and then $X_V \sim \Pr_V(\_|X_J)$. %
\end{Def}

\begin{Rem}
\begin{enumerate}
\item The definition \ref{def-cond-ind} assumes that the input variables $J$ are considered independent, in contrast to \cite{RERS17s,CD17s}, where all $J$ are implicitely assumed to be jointly confounded. We discuss this further in Supplementary Material \ref{confounded-input}.
\item In contrast with \cite{Daw79, CD17s, RERS17s} definition \ref{def-cond-ind} can accommodate any variable from $V$ or $J$ at any spot of the conditional independence statement.
\item  $\Indep_\Pr$ satisfies the \emph{separoid axioms} (see \cite{Daw79,Daw01,PeaPaz87,GeiVerPea90} or see rules 1-5 in Lem.\ \ref{GrAxPr} for $\Indep_\Pr$) as these rules %
are preserved under conjunction.
\end{enumerate}
\end{Rem}

\section{$\sigma$-SEPARATION}

In this section we will define $\sigma$-separation on directed mixed graphs (DMG) and present the generalized directed global Markov property stating that every \ioSCM\ will entail the conditional independencies that come from $\sigma$-separation in its induced DMG. We will again closely follow the work in \cite{FM18}.

\begin{Def}[Directed mixed graph (DMG)]
A \emph{directed mixed graph (DMG)} $G$ consists of a set of nodes $V$ together with a set of directed edges ($\tuh$) and bidirected edges ($\huh$).
In case $G$ contains no directed cycles it is called an \emph{acyclic directed mixed graph (ADMG)}.
\end{Def}

\begin{Def}[$\sigma$-Open walk in a DMG]
\label{s-open-walk}
Let $G$ be a DMG with set of nodes $V$ and $C \ins V$ a subset.
Consider a walk $\pi$ in $G$ with $n \ge 1$ nodes: 
\[ v_1 \dars  \cdots \dars v_n.\footnote{\label{fn2}The stacked edges are meant to be read as an ``OR'' at each place independently. We also allow for repeated nodes in the walks. Some authors also use the term ``path'' instead, which other authors use to refer to walks without repeated nodes.} \]
The walk will be called \emph{$C$-$\sigma$-open} if:
 \begin{enumerate}
   \item  
	the \emph{endnodes} $v_1, v_n \notin C$, and
   \item every triple of adjacent nodes in $\pi$ that is of the form:%
	\begin{enumerate}
	  \item \emph{collider}: $\qquad v_{i-1} \rars v_i \lars v_{i+1},$
		\item[] satisfies $v_i \in C$,
	 \item \emph{left chain}: $\qquad v_{i-1} \hut v_i \lars v_{i+1},$
	  \item[] satisfies  $v_i \notin C$ or $v_i \in C \cap \Sc^G(v_{i-1})$,
	 \item \emph{right chain}: $\qquad v_{i-1}\rars v_i \tuh v_{i+1},$
		\item[] satisfies $v_i \notin C$ or  $v_i \in C \cap \Sc^G(v_{i+1})$,
		\item \emph{fork}: $\qquad v_{i-1} \hut v_i \tuh v_{i+1},$
			 \item[] satisfies $v_i \notin C$ or 
			 \item[] $v_i \in C \cap \Sc^G(v_{i-1}) \cap \Sc^G(v_{i+1})$.
	\end{enumerate}
	\end{enumerate}
	
\end{Def}

Similar to d-separation we define $\sigma$-separation in a DMG.

\begin{Def}[$\sigma$-Separation in a DMG]
\label{s-sep-def}
Let $G$ be a DMG with set of nodes $V$. %
 Let $A,B,C \ins V$ be subsets. %
\begin{enumerate}
\item We say that $A$ and $B$ are \hyte{ns-sep}{$\sigma$-connected by $C$ or not $\sigma$-separated by $C$} if there exists a walk $\pi$ (with $n \ge 1$ nodes) in $G$ with one endnode in $A$ and one endnode in $B$ that is $C$-$\sigma$-open. In symbols this statement will be written as follows:
$$\qquad\displaystyle A \nIndep^\sigma_G B \given C.$$
\item Otherwise, we will say that $A$ and $B$ are \hyte{s-sep}{$\sigma$-separated} by $C$ and write:
$$\qquad\displaystyle A \Indep^\sigma_G B \given C.$$

\end{enumerate}
\end{Def}

\begin{Rem}
\begin{enumerate}

\item In any DMG we will always have that $\sigma$-separation implies d-separation, since every $C$-d-open walk is also $C$-$\sigma$-open because $\{v\} \ins \Sc^G(v)$.
\item If a DMG $G$ is acyclic, i.e.\ an ADMG, then $\sigma$-separation coincides with d-separation (also called m- or m$^*$-separation in this context).
\end{enumerate}
\end{Rem}

It was shown in \cite{FM17} %
that $\sigma$-separation satisfies the \emph{graphoid/separoid axioms} (see \cite{Daw79,Daw01,PeaPaz87,GeiVerPea90}):

\begin{Lem}[Graphoid and separoid axioms]
\label{GrAxPr}
Let $G$ be a DMG with set of nodes $V$ and $A,B,C,D \ins V$ subsets. Then we have the following rules for $\sigma$-separation in $G$ 
(with $\Indep$ standing for $\Indep^\sigma_G$):
\begin{enumerate}
\item Redundancy: $A \Indep B \given A$ always holds.
\item Symmetry: $ A \Indep B \given D \implies B \Indep A \given D$.
\item Decomposition:  $A \Indep B \cup C \given D$ 
$\implies A \Indep B \given D$.
\item Weak Union: $A \Indep B \cup C \given D$ 
$\implies A \Indep B \given C \cup D$.
\item Contraction: 
 $ (A \Indep B \given C \cup D) \land (A \Indep C \given D) $
\item[] $\qquad\implies A \Indep B\cup C \given D$.
\item Intersection: $ (A \Indep B \given C \cup D)\land(A \Indep C \given B \cup D)$%
\item[]$\qquad\implies A \Indep B \cup C \given D$, 
\item[] 
whenever $A,B,C,D$ are pairwise disjoint.
\item Composition: $(A \Indep B \given D) \land (A \Indep C \given D)$ 
\item[] $\qquad\implies A \Indep B\cup C \given D$.
\end{enumerate}
\end{Lem}

It was also shown that $\sigma$-separation is stable under \emph{marginalization} (see \cite{FM17,FM18}):

\begin{Thm}[$\sigma$-Separation under marginalization, see \cite{FM17,FM18}]
Let $G$ be a DMG with set of nodes $V$. Then for any sets $A,B,C \ins V$ and $L \ins V \sm (A \cup B \cup C)$ we have the equivalence:
\[ A \Indep^\sigma_{G} B \given C \;\iff\; A \Indep^\sigma_{G^{\sm L}} B \given C, \]
where $G^{\sm L}$ is the DMG that arises from $G$ by marginalizing out the variables from $L$.
\end{Thm}

\section{A GLOBAL MARKOV PROPERTY}

The most important ingredient for our results is a \emph{generalized directed global Markov property}  that relates the graphical structure of any \ioSCM\ $M$ to the conditional independencies of the observed random variables via a \emph{$\sigma$-separation criterion}.
Since we have no access to the latent nodes $u \in U$ of an \ioSCM\ with graph $G^+$ we need to marginalize them out (see Supplementary Material \ref{dmg}). 
This will give us an induced directed mixed graph (DMG) $G$.

\begin{Def}[Induced DMG of an \ioSCM]
\label{ind-s-CG}
Let $M=(G^+,\Xcal,\Pr_U,g)$ be an \ioSCM\ with $G^+=(V  \dot \cup U \dot \cup J, E^+)$.
The \emph{induced directed mixed graph (DMG) $G$ of $M$} is defined as follows:%
	 \begin{enumerate}
	   \item $G$ contains all nodes from $V \cup J$.
		 \item $G$ contains all the directed edges of $G^+$ whose endnodes are both in $V \cup J$.  
		 \item $G$ contains the bidirected edge $v \huh w$ with $v,w \in V$ if and only if $v \neq w$ and there exists a $u \in U$ with $v,w \in \Ch^{G^+}(u)$, i.e.\ $v$ and $w$ have a common latent confounder.
	 \end{enumerate}
	\end{Def}

The following \emph{generalized directed global Markov property} directly generalizes from mSCMs (see \cite{FM17,FM18}) to \ioSCMs. An alternative version with confounded input is given in \ref{input-confounde-mSCM-gdGMP-thm}.

\begin{Thm}[$\sigma$-Separation criterion]
\label{mSCM-gdGMP-thm} 
Let $M$ be an \ioSCM\ with induced DMG $G$. %
Then for all subsets $A,B,C \ins V \cup J$ we have the implication:
\[ A \Indep^\sigma_G B \given C \;\implies\; X_A \Indep_\Pr X_B \given X_C. \]
In words, if $A$ and $B$ are $\sigma$-separated by $C$ in $G$ then the corresponding variables $X_A$ and $X_B$ are conditionally independent given $X_C$ under $\Pr$, i.e.\ under the joint distribution $\Pr_U(X_V|\doit(X_J))\otimes\Pr_J(X_J)$ for \emph{any} product distribution $\Pr_J=\bigotimes_{j \in J} \Pr_j$.
\begin{proof}
As mentioned, after specifying the product distribution $\Pr_J$ the \ioSCM\ $M$ constitutes a well-defined mSCM with the same induced DMG $G$. So the $\sigma$-separation criterion for \ioSCMs\ directly follows from the mSCM-version proven in \cite{FM18,FM17}.
\end{proof}
\end{Thm}

\begin{Rem}
Note that, since $\sigma$-separation is stable under marginalization (see \cite{FM17,FM18}), also the $\sigma$-separation criterion is stable under marginalization.
\end{Rem}

\begin{Rem}[Causal calculus for mechanism change]
\label{soft-calc}
The $\sigma$-separation criterion \ref{mSCM-gdGMP-thm} can be viewed as the \emph{causal calculus for mechanism change} (also sometimes called ``soft'' interventions, see \cite{Pearl09,MMC18s,MGS05s,EM07s}).
As an example consider $A,B \ins V$, $I \ins J$. %
Then the graphical separation $A \Indep^\sigma_G I \given B \cup (J \sm I)$
implies that the conditional probability $\Pr_U(X_A | X_B ,\doit(X_J))$ is independent of the actual input variables in $I$.
\end{Rem}

\section{THE EXTENDED IOSCM}

In this section we want to consider (perfect) interventions onto the observed nodes and improve upon the general rules mentioned in \ref{soft-calc}.
For an elegant treatment of this we need to gather for a given \ioSCM\ $M$ \emph{all} interventional \ioSCMs\ $M_{\doit(W)}$, where $W$ runs through all subsets of observed variables, and glue them all together into one big \emph{extended \ioSCM} $\hat M$. 
To consider all interventions at once we will need to introduce additional intervention variables $I_v$ to the graph $G^+$, $v \in V$, which indicate which interventional mechanisms to use.
Such techniques were already used in the acyclic case in \cite{Pearl93as,Pearl93bs,Pearl09}. The definition will be made in such a way that $\hat M$ will still be a well-defined \ioSCM. So all the results for \ioSCMs\ will apply to $\hat M$, most importantly the $\sigma$-separation criterion (Thm.\ \ref{mSCM-gdGMP-thm}).

\begin{Def}
\label{ext-mSCM}
Let $M=(G^+,\Xcal,\Pr_U,g)$ be an \ioSCM\ with $G^+=(V \dot \cup U \dot \cup J, E^+)$. 
The \emph{extended \ioSCM} $\hat M = (\hat G^+, \hat \Xcal, \Pr_U, \hat g)$ will be defined as follows:
\begin{enumerate}
	\item For every $v \in V$ define the interventional domain $\Ical_v:=\Xcal_v \dot \cup \{\obs_v\}$, where $\obs_v$ is a new symbol corresponding to the observational (non-interventional) regime. For a set $A \ins V$ we put $\Ical_A:=\prod_{v \in A} \Ical_v$ and $\obs_A:=(\obs_v)_{v \in A}$.
	\item Let $\hat G^+$ be the graph $G^+$ with the additional intervention nodes $I_v$ and directed edges 
	$I_v \tuh v$ for every $v \in V$. For a uniform notation we sometimes write $I_j$ instead of $j$ for $j \in J$. So we have:
	\item[]  $\hat J:= J \cup \{ I_v \,|\, v \in V\} = \{ I_w \,|\, w \in V \dot\cup J\}$.
	 \item For every $A \ins V$ we will define the mechanism:
	$$ \hat g_A :\; \hat \Xcal_{\Pa^{\hat G^+}(A)\sm A}= \Ical_A \times \Xcal_{\Pa^{G^+}(A)\sm A}  \to  \Xcal_A = \hat \Xcal_A. $$
	First, for $x_A \in \Ical_A$ we put $I(x_A):=\{ v \in A | x_v \neq \obs_v \}$.	
	Consider the subgraph of $G^+$:
	$$H(x_A):=(\Pa^{G^+}(A)\cup A)_{\doit(I(x_A))}.$$
	Then define recursively for $v \in A$:
	$$ \hat g_{A,v}(x_A,x_{\Pa^{G^+}(A)\sm A}) $$
	$$:=\left\{ 
	\begin{array}{lcl}
	 x_v  &\text{ if }&  v \in I(x_A),\\
	 g_{S,v}(x_{\Pa^{H(x_A)}(S)\sm S}) &\text{ if }&  v \notin I(x_A),
	\end{array}\right.
	$$
	where $S:=\Sc^{H(x_A)}(v)$ is also a loop in $G^+$.
	\item These functions then are again globally compatible and $\hat M$ constitutes a well-defined \ioSCM.
	\item All the distributions in $\hat M$ then are given by the general procedure of \ioSCMs\ (see Def.\ \ref{iSCM-variables}). 
	 We introduce the notation for $C \ins V$ and $(x_C,x_J) \in \Ical_C \times \Xcal_J$:
	 \begin{eqnarray*}
	&&\Pr_U(X_V|I_C=x_C,X_J=x_J)\,:=\\ 
	&&  \Pr_U(X_V|\doit((I_C,I_{V\sm C},X_J)=(x_C,\obs_{V\sm C},x_J)).
	\end{eqnarray*}
	\item The \emph{extended DMG} $\hat G$ of $G^+$ is then the induced DMG of $\hat G^+$, i.e.\ the induced DMG $G$ with the additional edges $I_v \tuh v$ for every $v \in V$.
\end{enumerate}
\end{Def}

The following result now relates the interventional distributions of the \ioSCM\ $M$ with the ones from the extended \ioSCM\ $\hat M$. These relations will be used in the following.

\begin{Prp}
\label{caus-calc-eqn}
Let $M=(G^+,\Xcal,\Pr_U,g)$ be an \ioSCM\ with $G^+=(V \dot \cup U \dot \cup J, E^+)$ and $\hat M$ the extended \ioSCM.
Let $A,B,C \ins V$ 
be pairwise disjoint set of nodes and $x_{C\cup J} \in \Xcal_{C \cup J}$. Then we have the equations:
\begin{align*} 
 &\; \Pr_U(X_A|X_B, \doit(X_{C\cup J}=x_{C \cup J})) \\
=&\; \Pr_U (X_A | X_B, I_C=x_C,X_J=x_J) \\
 =&\; \Pr_U (X_A | X_B, I_C=x_C, X_C=x_C,X_J=x_J).
\end{align*}
\begin{proof} This follows from $I(x_C,\obs_{V \sm C})=C$. See Supplementary Material \ref{caus-calc-eqn-proof}. %
\end{proof}
\end{Prp}
\section{THE THREE MAIN RULES OF CAUSAL CALCULUS}

\begin{Not}
Since everything has been defined in detail in the last section we now want to make use of a simplified and more suggestive notation for better readability.
\begin{enumerate}
	\item We identify variables $X_A$ with the set of nodes $A$.
	\item We omit values $x_V$ and the subscript in $\Pr_U$. E.g.\ we write
	 $\Pr(Y|I_T,T,Z, \doit(W))$ instead of 
    \begin{equation*}\begin{split}\Pr_U\big(X_Y\,|\,&I_T=x_T,X_T=x_T,X_Z=x_Z,\\
    &\doit(X_W=x_W)\big),\end{split}\end{equation*}
	where the latter comes from the extended \ioSCM\ of the intervened \ioSCM\ $M_{\doit(W)}:=M_{\doit(W \sm J)}$ of $M$.
	\item We abbreviate $X_Y \Indep_{ \Pr_U(\_|\doit(X_W=x_W))}  X_T |X_Z$ as
    $Y \Indep_\Pr T |Z, \doit(W)$, etc..
	\item We write $Y \Indep_G^\sigma I_X \given X, Z, \doit(W)$ to mean $Y \Indep_{\hat G_{\doit(W)}}^\sigma I_X \given X, Z$, where $\hat G_{\doit(W)}$ is the extended DMG of the intervened graph $ G^+_{\doit(W)}$.
\end{enumerate}
\end{Not}

\begin{Thm}[The three main rules of causal calculus]
\label{three-rules}
Let $M$ be an \ioSCM\ with set of observed nodes $V$ and input nodes $J$ and induced DMG $G$.
Let $X,Y,Z \ins V$ and $J \ins W \ins V \cup J$ be subsets.
\begin{enumerate}
\item Insertion/deletion of observation: 
\item[] If $\qquad\displaystyle Y \Indep_{ G }^\sigma X |Z, \doit(W)\qquad$ then:
 $$\Pr(Y|X, Z, \doit(W)) = \Pr(Y| Z, \doit(W)).$$
\item Action/observation exchange:
\item[] If $\qquad\displaystyle Y \Indep_{G}^\sigma I_X | X, Z, \doit(W)\qquad$ then:
$$\Pr(Y| \doit(X), Z,  \doit(W)) = \Pr(Y|X, Z,  \doit(W)).$$
\item Insertion/deletion of actions:
\item[] If $\qquad\displaystyle Y \Indep_{G}^\sigma I_X | Z,  \doit(W)\qquad$ then:
$$\Pr(Y| \doit(X), Z,  \doit(W)) = \Pr(Y| Z, \doit(W)).$$
\end{enumerate}
\end{Thm}

The proofs follow directly from the $\sigma$-separation criterion \ref{mSCM-gdGMP-thm} and Prp. \ref{caus-calc-eqn} applied to the extended \ioSCM\ and can be found in Supplementary Material \ref{three-rules-proofs}.

\section{ADJUSTMENT CRITERIA}

\begin{Not}
\label{selection-backdoor-not}
Let $M=(G^+,\Xcal,\Pr,g)$ be an \ioSCM\ with $G^+=(V \dot \cup U \dot \cup J, E^+)$. 
The following set of nodes/variables will play the described roles:\\[0.5\baselineskip]
  \centerline{\begin{tabular}{ll}
  $Y$: & the outcome variables,\\
  $X$: & the treatment or intervention variables,\\
  $Z_0$: & the core set of  adjustment variables,\\
  $Z_+$: & additional adjustment variables,\\
  \multicolumn{2}{l}{$Z := Z_0 \cup Z_+$: all actual adjustment variables,}\\
  $L$: & ``marginalizable'' adjustment variables,\\
  $C$: & context variables,\\
  $W$: & default intervention variables containing $J$,\\
  $S$: & variables inducing selection bias given $S=s$.
  \end{tabular}}
\end{Not}

We are interested in finding a ``$\doit(X)$-free'' expression for the (conditional) causal effect $\Pr(Y|C,\doit(X),\doit(W))$ only using  data for $C,X,Y,Z$ that was gathered under selection bias $S=s$ and intervention $\doit(W)$  and additional unbiased observational data for $C,Z$ given 
$\doit(W)$. The task can be achieved via the following criterion, which is a generalization of the acyclic case  of the selection-backdoor criterion (see \cite{BTP14s}), the backdoor criterion (see \cite{Pearl93as,Pearl93bs,Pearl09}) and its extensions (also see \cite{PearlPaz13,SWR10s,PTKM15s,CB17s}) to general \ioSCMs.

\begin{Thm}[General adjustment criterion and formula]
\label{adjustment}
Let the setting be like in \ref{selection-backdoor-not}.
Assume that data was collected under selection bias, $\Pr(V|S=s,\doit(W))$ (or under $\Pr(V|\doit(W))$ and $S=\emptyset$), and there are unbiased samples from $\Pr(Z|C,\doit(W))$.
Further assume that the variables satisfy:
\begin{enumerate}
\item $\displaystyle (Z_0,L) \Indep^\sigma_{G} I_X \given C,\doit(W)$, and
\item $\displaystyle Y \Indep^\sigma_{G} (I_X, Z_+) \given C, X, Z_0,L,\doit(W)$, and
\item $\displaystyle Y \Indep^\sigma_{G} S \given C,X, Z,\doit(W)$, and 
\item $\displaystyle L \Indep^\sigma_{G} X \given C,Z,\doit(W)$.
\end{enumerate}
Then one can estimate the conditional causal effect $\Pr(Y|C,\doit(X),\doit(W))$ via the \emph{adjustment formula}:
\begin{eqnarray*}
&& \Pr(Y|C,\doit(X),\doit(W)) \\
&=& \int \Pr(Y | X,Z, C, S=s,\doit(W))\, d\Pr(Z|C,\doit(W)).
\end{eqnarray*}
\end{Thm}

The proof again follows directly from the $\sigma$-separation criterion \ref{mSCM-gdGMP-thm} and Prp. \ref{caus-calc-eqn} applied to the extended \ioSCM\ and can be found in the Supplementary Material \ref{adjustment-proofs}.

\begin{figure}[h]
\centering
\begin{tikzpicture}[scale=.7, transform shape]
\tikzstyle{every node} = [draw,shape=circle,color=blue]
\node[shape=rectangle, color=teal] (IX) at (0,0) {$I_X$};
\node (X) at (2,0) {$X$};
\node (W2) at (4,1) {$W$};
\node (Y) at (7,0) {$Y$};
\node (Z0) at (2.5,2) {$Z_0$};
\node (L1) at (5.5,2) {$L_1$};
\node[shade] (C) at (0,2.1) {$C$};
\node (Z1) at (2,-1.7) {$Z_1$};
\node (Z2) at (4,-2) {$Z_2$};
\node[blue] (L2) at (6.2,-1.7) {$L_2$}; %
\node[shade] (S) at (0,-2) {$S$};
\draw[-{Latex[length=3mm, width=2mm]}, color=teal] (IX) to (X);
\draw[-{Latex[length=3mm, width=2mm]}, color=blue] (X) to (Z1);
\draw[-{Latex[length=3mm, width=2mm]}, color=blue] (X) to (Y);
\draw[-{Latex[length=3mm, width=2mm]}, color=blue] (Z1) to (Z2);
\draw[-{Latex[length=3mm, width=2mm]}, color=blue] (Z1) to (Z2);
\draw[-{Latex[length=3mm, width=2mm]}, color=blue] (C) to (Z0);
\draw[-{Latex[length=3mm, width=2mm]}, color=blue] (Z0) to (X);
\draw[-{Latex[length=3mm, width=2mm]}, color=blue, bend left] (Z0) to (L1);
\draw[-{Latex[length=3mm, width=2mm]}, color=blue, bend left] (L1) to (W2);
\draw[-{Latex[length=3mm, width=2mm]}, color=blue, bend left] (W2) to (Z0);
\draw[{Latex[length=3mm, width=2mm]}-{Latex[length=3mm, width=2mm]}, color=red, bend left] (L1) to (Y);
\draw[-{Latex[length=3mm, width=2mm]}, color=blue] (Z1) to (S);
\draw[-{Latex[length=3mm, width=2mm]}, color=blue] (L2) to (Y); %
\draw[-{Latex[length=3mm, width=2mm]}, color=blue] (L2) to (Z2); %
\end{tikzpicture}
  \caption{An induced DMG $G$ with input node $I_X$ (the others are left out for readability). The variables satisfy the general adjustment criterion for $\Pr(Y|C,\doit(X))$ with $L=\{L_1,L_2\}$ and $Z_+=\{Z_1,Z_2\}$. Note that $L_2$ could also have been a latent variable. Different colours for different node and/or edge types. 
	}
 \label{fig:mSCM-DMG}
\end{figure}
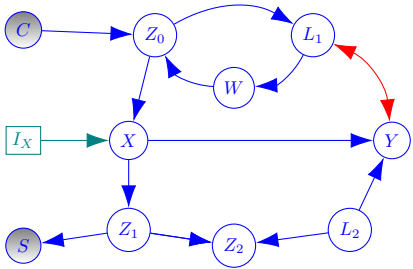

\begin{Rem}
Note that the adjustment \emph{formula} in theorem \ref{adjustment} does not depend on $L$. This thus allows us to even choose variables for $L$ that come from an \ioSCM\ $M'$ that marginalizes to $M$, e.g.\ $L \ins U$ or by extending directed edges $v \tuh w$ by $v \tuh \ell \tuh w$ with $\ell \in L$. This technique was used in \cite{SWR10s} to find \emph{all} adjustment sets in the \emph{acyclic} case with $C=S=\emptyset$. %
\end{Rem}

\begin{Cor}
\label{backdoor-cor}
Let the notations be like in \ref{selection-backdoor-not} and \ref{adjustment} and $W=J=\emptyset$.
We have the following special cases, which in the \emph{acyclic} case will reduce to the ones given by the indicated references:
\begin{enumerate}
\item General selection-backdoor (see \cite{CB17s}): $C=\emptyset$.
\item Selection-backdoor (see \cite{BTP14s}): $C=L=\emptyset$. 

\item Extended backdoor %
 (see \cite{PearlPaz13,SWR10s}): $C=S=\emptyset$. %
\item Backdoor (see \cite{Pearl93as,Pearl93bs,Pearl09}): $C\!=\! S\!=\! L\!=\! Z_+\!=\!\emptyset$:%
\begin{enumerate}
\item $\displaystyle Z \Indep^\sigma_{G} I_X$, and %
\item $\displaystyle Y \Indep^\sigma_{G} I_X \given X, Z$,  implies:
\end{enumerate}
$$\displaystyle \Pr(Y|\doit(X)) = \int \Pr(Y|X,Z)\, d\Pr(Z).$$
\end{enumerate}
\end{Cor}

More details can be found in the Supplementary Material \ref{backdoor-cor-supp}.
Also a generalization of the criterion for selection without/partial external data of \cite{CB17s,CTB18s} is given there.

\begin{Rem}
The conditions in theorems \ref{three-rules}, \ref{adjustment} and corollary \ref{backdoor-cor} are in the acyclic setting usually phrased in terms of sub-structures of the graph $G$ (see \cite{Pearl93as,Pearl93bs,Pearl09}):
\begin{enumerate}
\item For rule 3 in Thm. \ref{three-rules} one usually requires $Y \Indep^d X | Z,  W$ in the graph $G_{\doit(W)}$ that is further mutilated on the set $X(Z)$, the set of all $X$-nodes that are not ancestors of any $Z$-node in $G_{\doit(W)}$.
\item For the backdoor criterion instead of $L \Indep^d_G I_X$ we could have written that $L$ does not contain any descendent of $X$; and for 
$Y \Indep^d_G I_X \given X, Z$ that $Z$ blocks all ``backdoor paths'' from $X$ to $Y$.
\end{enumerate}
We presented the results in the formulaic terms of $\sigma$-separation because the relations to their use is directly indicated (e.g.\ in the proofs), it makes the generalization to \ioSCMs\ possible and when reduced to the acyclic case it will be equivalent to the usual description.
\end{Rem}

\section{IDENTIFYING CAUSAL EFFECTS}

Here we extend the \emph{ID algorithm} for the identification of causal effects to \ioSCMs.
The main references are \cite{Pearl09,GP95s,Tian2002,TP02s,Tian04s,SP06,HV06s,HV08s,RERS17s}.
The task is to decide if a causal effect $\Pr(Y|\doit(W))$ in an \ioSCM\  
can be \emph{identified} from (i.e., expressed in terms of) the observational distributions $\Pr(V|\doit(J))$
and the induced graph $G$.
Note that having more dependence structure (like latent confounders, feedback cycles, etc.) will leave us with less identifiable causal effects in general.
Due to space limitations, we can only provide here the bare necessities to state the
generalized ID algorithm. We assume that the reader is already familiar with the
ID algorithm formulated for ADMGs (for example, the treatment in \cite{Tian04s}).

We generalize the notion of districts / C-components:
\begin{Def}[Consolidated districts]
Let $G$ be a directed mixed graph (DMG) with set of nodes $V$. Let $v \in V$.
The \emph{consolidated district} $\CDist^G(v)$ of $v$ in $G$ is given by all nodes $w \in V$ for which there exist $k \ge 1$ nodes 
$(v_1,\dots,v_k)$ in $G$ such that $v_1=v$, $v_k=w$ and for $i=2,\dots, k$ we have that the bidirected edge $v_{i-1} \huh v_i$ is in $G$ or that $v_{i} \in \Sc^G(v_{i-1})$.
  For $B \ins V$ we write $\CDist^G(B) := \bigcup_{v\in B} \CDist^G(v)$.
  Let $\CDcal(G)$ be the set of consolidated districts of $G$. 
\end{Def}

We also generalize the notion of topological order:
\begin{Def}[Apt-order, see \cite{FM17}]
Let $G$ be a DMG with set of nodes $V$.
An \emph{assembling pseudo-topological order (apt-order)} of $G$ is a total order $<$ on $V$ with the following two properties:
\begin{enumerate}
\item For every $v,w \in V$ we have: 
$$w \in \Anc^G(v)\sm \Sc^G(v) \, \implies \, w < v.$$
\item For every $v_1,v_2,w \in V$ we have:
$$ v_2 \in \Sc^G(v_1) \land (v_1 \le w \le  v_2) \,\implies\, w \in \Sc^G(v_1).$$
\end{enumerate}
\end{Def}
 
\begin{Rem}
Let $G$ be a DMG.
\begin{enumerate}
\item If $G$ is acyclic then an apt-order $<$ is the same as a topological order (i.e.\ $w \in \Pa^G(v) \implies w < v$).
\item If $G$ has a topological order then $G$ is acyclic.
\item For any DMG $G$ there always exists an apt-order $<$ (in contrast to topological orders).
\end{enumerate}
\end{Rem}

\begin{Not}
Let $G$ be a DMG with set of nodes $V$ and $<$ a apt-order on $G$. 
For elements $v\in V$ and subsets $B \ins V$ we put:
\begin{enumerate}
	\item $\Pred^G_<(v):= \{ w \in V \,|\, w <v \}$,
	\item $\Pred^G_\le(v):= \{ w \in V \,|\, w =v \text{ or } w <v \}$,
	\item $\Pred^G_\le(B) := \bigcup_{v \in B} \Pred^G_\le(v)$,
	\item $\Pred^G_<(B) := \Pred^G_\le(B) \sm B$.
\end{enumerate}
\end{Not}
\begin{Rem}
If $B$ is strongly-connected, then $\Pred^G_\le(B)$ is ancestral in $G$,
i.e., $\Anc^G(\Pred^G_\le(B)) = \Pred^G_\le(B)$.
\end{Rem}

The notion of input variables enables the following convenient and intuitive construction:
\begin{Def}[Sub-\ioSCMs]
\label{sub-iSCM}
Let $M=(G^+,\Xcal,\Pr_U,g)$ be an \ioSCM\ with $G^+=(V \dot \cup U \dot \cup J, E^+)$. 
For $C \ins V$ non-empty define the \ioSCM\ $M_{[C]}$ as follows:
\begin{enumerate}
\item Put $G^+_{[C]}$ to be the subgraph of $G^+_{\doit(\Pa^{G}(C) \sm C)}$ %
    induced by $C \cup \Pa^{G^+}(C)$.
\item $V_{[C]}:=C$, $J_{[C]}:= \Pa^{G^+}(C)\sm (C \cup U)$, $U_{[C]}:= U \cap \Pa^{G^+}(C)$.
\item Keep all functions $g_S$ with $S \ins C$.
\item $\Pr_{U_{[C]}}:=\bigotimes_{u\in U_{[C]}} \Pr_u$, i.e.\ the marginal of $\Pr_U$ and we will use the notation $\Pr_U$ (or just $\Pr$) for both.
\end{enumerate}
For $C \ins V \cup J$ with $C \cap V \neq \emptyset$ put %
  $M_{[C]}:=M_{[C \cap V]}$.
\end{Def}

By the definition of the random variables induced by an \ioSCM\ we immediately get the following basic result:
\begin{Lem}\label{sub-iSCM-remark}
Let $M=(G^+,\Xcal,\Pr_U,g)$ be an \ioSCM\ with $G^+=(V \dot \cup U \dot \cup J, E^+)$. 
For $C \ins V$, we have (indices for emphasis):
$$\Pr_{M_{[C]}}(C|\doit(\Pa^G(C) \sm C)) = \Pr_M(C|\doit(J \cup W)),$$
for any $W \ins V \sm C$ that contains $(\Pa^G(C) \cap V)\sm C$.
As a special case: if $A \ins G$ is ancestral, i.e., $\Anc^G(A) = A$,
$$\Pr_{M_{[A]}}(A \cap V|\doit(A \cap J)) = \Pr_M(A \cap V|\doit(J \cup W))$$
for any $W \ins V \sm A\cap V $.
\end{Lem}

The ID algorithm works by repeatedly applying the previous lemma and the following rules:
\begin{Prp}
\label{id-eqn}
Let $M=(G^+,\Xcal,\Pr_U,g)$ be an \ioSCM\ with $G^+=(V \dot \cup U \dot \cup J, E^+)$ and 
$<$ an apt-order for $G^+$. %
\begin{enumerate}
\item $$\Pr(V|\doit(J)) = \bigotimes_{\substack{S \in \Scal(G)\\S \ins V}} \Pr(S |  \Pred^G_<(S) \cap V, \doit(J)).$$
\item 
For $S \ins V$ a strongly connected component of $G$, $D:=\CDist^G(S)$ its consolidated district in $G$ and $P:=\Pa^G(D)\sm D$:
\begin{align*}
  & \Pr_M(S|\Pred_<^G(S)\cap V,\doit(J)) \\
  &= \Pr_{M_{[D]}}(S|\Pred_<^{G_{[D]}}(S) \cap D,\doit(P)).
\end{align*}

\item For $D \ins V$ a consolidated district of $G$:
$$\begin{array}{rcl}
&&\Pr(D|\doit(J \cup V \sm D)) \\
  &=& \displaystyle\bigotimes_{\substack{S \in \Scal(G)\\S \ins D}} \Pr(S | \Pred^G_<(S) \cap V, \doit(J)).
\end{array}$$
\end{enumerate}
\begin{proof} %
1.\ uses the chain rule; 2.\ is proved in Supplementary Material \ref{new-central-lemma}; 3.\ is shown by applying 1.\ and Remark \ref{sub-iSCM-remark} to $G_{[D]}$ and then making use of 2..
\end{proof}
\end{Prp}

\begin{Rem}
\label{order-rem}
Naively putting the equations of Prp. \ref{id-eqn} into each other would give us the equation:
$$\Pr(V|\doit(J)) = \bigg[\bigotimes_{\substack{D \in \CDcal(G)\\D \ins V}}\bigg] \Pr(D|\doit(J \cup V \sm D)).$$
Note that the product might not be well-defined 
as the consolidated districts i.g.\ are not totally ordered by $<$ (in contrast to strongly connected components), even in the acyclic case. For example, consider the graph:
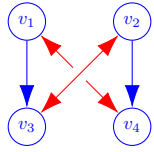
\[
\begin{tikzpicture}[scale=.7, transform shape]
\node[shape=rectangle, color=transparent] (m) at (1,1) {};
\tikzstyle{every node} = [draw,shape=circle,color=blue]
\node (v1) at (0,2) {$v_1$};
\node (v2) at (2,2) {$v_2$};
\node (v3) at (0,0) {$v_3$};
\node (v4) at (2,0) {$v_4$};
\draw[-{Latex[length=3mm, width=2mm]}, color=blue] (v1) to (v3);
\draw[-{Latex[length=3mm, width=2mm]}, color=blue] (v2) to (v4);
\draw[-{Latex[length=3mm, width=2mm]}, color=red] (m) to (v1);
\draw[-{Latex[length=3mm, width=2mm]}, color=red] (m) to (v4);
\draw[{Latex[length=3mm, width=2mm]}-{Latex[length=3mm, width=2mm]}, color=red] (v2) to (v3); %
\end{tikzpicture}
\]
This problem is usually not addressed in the literature. The problem disappears if every strongly connected component $S \ins V$ comes with a measure $\mu_S$ such that 
$\Pr(V|\doit(J))$ has a density w.r.t.\ the product measure $\bigotimes_{\substack{S \in \Scal(G)\\S \ins V}} \mu_S$.
Then the densities $p(D|\doit(J \cup V \sm D))$ can be multiplied in any order and the integration can be separately done via the  $\mu_S$ in reverse order of $<$. 
\end{Rem}

We now have all the prerequisites to state the generalized ID algorithm (Algorithm~\ref{ID}) and prove its correctness:
\begin{Thm}[Consequence of \ref{id-eqn}, \ref{order-rem}]
\label{ID-theorem}
Let $M=(G^+,\Xcal,\Pr_U,g)$ be an \ioSCM\ with $G^+=(V \dot \cup U \dot \cup J, E^+)$
 with set of observed nodes $V$ and input nodes $J$ and distributions $\Pr(V|\doit(J))$.
Let $<$ be an apt-order for $G^+$.
Assume that for every strongly connected component $S \ins V$ we have a measure $\mu_S$ such that 
$\Pr(V|\doit(J))$ has a density w.r.t.\ the product measure $\bigotimes_{\substack{S \in \Scal(G)\\S \ins V}} \mu_S$.
Let $Y \ins V$ and $W \ins J \cup V$ be subsets.
If the extended ID algorithm (see Algorithm~\ref{ID}) does not ``FAIL'' then the causal effect 
$\Pr(Y|\doit(W))$ is identifiable, i.e.\ it can be computed from $\Pr(V|\doit(J))$ alone, and the expression is obtained by postprocessing the output of the algorithm.
\end{Thm}

\begin{algorithm}[t]
\caption{
  \textsc{Id}: Generalized ID algorithm for the identification of causal effects in general \ioSCMs.}
\label{ID}
\begin{algorithmic}[1]
  \Function{Id}{$G,Y,W,\Pr(V|\doit(J))$}
  \State \textbf{require:} $Y \ins V$, $W \ins V$, $Y \cap W = \emptyset$
  \State $H \gets \Anc^{G_{V\sm W}}(Y)$
  \For{$C \in \CDcal(H)$}
  \State $Q[C] \gets \text{\textsc{IdCD}}(G,C,\CDist^G(C),Q[\CDist^G(C)])$
  \If{$Q[C] = \text{FAIL}$}
    \State \textbf{return} FAIL
  \EndIf
  \EndFor
  \State $Q[H] \gets \left[\bigotimes_{C \in \CDcal(H)}\right] Q[C]$
  \State \textbf{return} $\Pr(Y|\doit(J, W)) = \int Q[H] dx_{H \sm Y}$
  \EndFunction
  \Statex
  \Function{IdCD}{$G,C,D,Q[D]$}
  \State \textbf{require:} $C \ins D \ins V$, $\CDcal(G_D)=\{D\}$
  \State $A \gets \Anc^{G_{[D]}}(C) \cap D$ 
  \State $Q[A] \gets \int Q[D] d(x_{D\sm A})$
  \If{$A=C$}
    \State \textbf{return} $Q[A]$
  \ElsIf{$A=D$}
    \State \textbf{return} FAIL
  \ElsIf{$C \subsetneq A \subsetneq D$}
    \For{$S \in \Scal(G_{[A]})$ s.t.\  $S \ins \CDist^{G_{[A]}}(C)$}
    \State $R_A[S] \gets \Pr(S | \Pred^{G}_<(S) \cap A, \doit(J \cup V \sm A))$
    \EndFor
    \State $Q[\CDist^{G_{[A]}}(C)] 
    \gets \bigotimes_{\substack{S \in \Scal(G_{[A]})\\S \ins \CDist^{G_{[A]}}(C)}} R_A[S]$
    \State \textbf{return} 
      \textsc{IdCD}$(G,C,\CDist^{G_{[A]}}(C),Q[\CDist^{G_{[A]}}(C)])$
  \EndIf
  \EndFunction
\end{algorithmic}
\end{algorithm}

\begin{Rem}
\label{rem-id}
\begin{enumerate}
\item We make no claim about the completeness of the algorithm here.
\item The algorithm reduces to the usual version in the acyclic case (see \cite{Tian2002,TP02s,Tian04s,RERS17s}).
\item The main idea of the generalized ID algorithm is to exploit that the causal effects onto ancestral subsets and consolidated districts are identifiable. The algorithm then alternates these constructions to shrink towards the queried set $C$ until convergence, i.e.\ until a set $A$ is reached that is both the ancestral closure of $C$ and a consolidated district in itself. If $C=A$ then the causal effect onto $C$ is identifiable, otherwise it outputs ``FAIL'' as no shrinking can be done with these techniques anymore.
Also see Supplementary Material \ref{id-rem-supp}. 

\end{enumerate}
\end{Rem}

\section{CONCLUSION}    %

We proved the three main rules of causal calculus and general adjustment criteria with corresponding formulas to recover from interventions and selection bias  for general \ioSCMs, which allow for arbitrary probability distributions, non-/linear functional relations, latent confounders, external non-/probabilistic parameter/action/intervention/context/input nodes and \emph{cycles}. This generalizes all the corresponding results of \emph{acyclic} causal models (see \cite{Pearl93as,Pearl93bs,Pearl09,PearlPaz13,SWR10s,BTP14s,PTKM15s,CB17s}) 
to general \ioSCMs. We also showed how to extend the ID algorithm for the identification of causal effects from the acyclic setting to general \ioSCMs.
In supplementary material \ref{twin} we also show how to do counterfactual reasoning in \ioSCMs.
Future work might address completeness questions of the ID algorithm (see \cite{SP06,SP06s,HV06s,Pearl09}).

\subsubsection*{Acknowledgements}
This work was supported by the European Research Council (ERC) under the European Union's Horizon 2020 research and innovation programme (grant agreement 639466).

\newpage
\bibliographystyle{plain}
\small

\providecommand{\bysame}{\leavevmode\hbox to3em{\hrulefill}\thinspace}
\providecommand{\MR}{\relax\ifhmode\unskip\space\fi MR }
\providecommand{\MRhref}[2]{%
  \href{http://www.ams.org/mathscinet-getitem?mr=#1}{#2}
}
\providecommand{\href}[2]{#2}

\clearpage
\appendix
\normalsize

{\Large{\bf SUPPLEMENTARY MATERIAL}}

\section{TWIN NETWORKS AND COUNTERFACTUALS}
\label{twin}
In addition to probabilistic and causal reasoning about interventions, \ioSCMs\ allow for counterfactual reasoning.
Given an \ioSCM\ $M$ with graph $G^+=(V \dot \cup U \dot \cup J, E^+)$, a set $W \ins V \cup J$  and the corresponding intervened \ioSCM\ $M_{\doit(W)}$ with graph $G^+_{\doit(W)}$  one can construct a (merged) \emph{twin \ioSCM} $M_\text{twin}$  similarly to the acyclic case (see \cite{Pearl09}), or a single world intervention graph (SWIG, see \cite{SWIG13s}). This is done by identifying/merging the corresponding nodes, mechanisms and variables from the non-descendants of $W$, i.e., $\NonDesc^{G^+}(W)$ and 
$\NonDesc^{G^+_{\doit(W)}}(W)$, which are unchanged by the action $\doit(W)$. Then one has the two different branches $\Desc^{G^+}(W)$ and $\Desc^{G^+_{\doit(W)}}(W)$ in the network. This construction then allows one to formulate counterfactual statements like in the acyclic case (see \cite{Pearl09}), but now for general \ioSCMs. E.g., one could state the assumption of \emph{strong ignorability} (see \cite{RR83,Pearl09})  as:
$$ \left( Y^{\doit(\obs)}, Y^{\doit(X)} \right) \Indep^\sigma_{G_\text{twin}} X \given Z, $$
or the \emph{conditional ignorability} (see \cite{RR83,SWR10s}) as:
$$ Y^{\doit(X)} \Indep^\sigma_{G_\text{twin}} X \given Z.$$
All the causal reasoning rules derived in this paper can thus also be applied to reason about counterfactuals.%

\section{MARGINALIZATION OF DIRECTED MIXED GRAPHS}
\label{dmg}

For completeness, we provide here the definition of marginalization of directed mixed graph. For more details and the relationship with the marginalization of an mSCM (or as a straightforward generalization, an ioSCM), we refer the reader to \cite{FM17}.
\begin{Def}[Marginalization of DMGs]
Let $G=(V,E,B)$ be a directed mixed graph (DMG) with set of nodes $V$, directed edges $E$ and bidirected edges $B$.
Let $W \ins V$ be a subset of nodes.
We define the \emph{marginalized DMG} $G^{\sm W}:=G'=(V',E',B')$ (``marginalizing out $W$''), also called \emph{latent projection of $G$ onto $V \sm W$}, with set of nodes $V':=V \sm W$ via the following rules (for $v_1,v_2 \in V \sm W=V'$):
\begin{enumerate}
\item $v_1 \tuh v_2 \in E'$ iff there exist $k \ge 0$ nodes $w_1,\dots,w_k \in W$ such that the directed walk: 
$$v_1 \tuh w_1 \tuh \cdots \tuh w_k \tuh v_2$$
 lies in $G$ (the corner case $v_1 \tuh v_2 \in E$ also applies).
\item $v_1 \huh v_2 \in B'$ iff there exist $k \ge 0$ nodes $w_1,\dots,w_k \in W$ and an index $0 \le m \le k$ such that a walk of the form:
$$v_1 \hut w_1 \hut \cdots \hut w_m \tuh\cdots \tuh w_k \tuh v_2$$ 
lies in $G$ with $m \ge 1$ or a walk of the form: 
$$v_1  \underbrace{\hut w_1 \hut \cdots \hut w_{m}}_{m \ge 0} \huh \underbrace{w_{m+1} \tuh \cdots \tuh w_k \tuh }_{k-m \ge 0} v_2$$
 lies in $G$ (including the corner cases $v_1 \huh v_2 \in B$ and  $v_1 \hut w \tuh v_2$ in $G$ with $w \in W$).
\end{enumerate}
\end{Def}

\section{CONDITIONAL INDEPENDENCE AND ITS ALTERNATIVE WITH CONFOUNDED INPUTS}
\label{confounded-input}

Here we want to give a generalization of \cite{RERS17s,CD17s} in the flavor of definition \ref{def-cond-ind}.
The main point is that the approaches of conditional independence for families of distributions/Markov kernels in \cite{RERS17s,CD17s} implicitely assume that the input variables $J$ are jointly confounded. The definition \ref{def-cond-ind} of conditional independence, in contrast, assumes (via the product distributions) that the variables $J$ are jointly independent. The approach in definition \ref{def-cond-ind} can be easily adapted to the confounded input setting as follows.

\subsection{INPUT CONFOUNDED CONDITIONAL INDEPENDENCE}

\begin{Def}[Input confounded conditional independence]
\label{def-input-confounded-cond-ind}
Let $\Xcal_V:=\prod_{v \in V} \Xcal_v$ and $\Xcal_J:=\prod_{j \in J} \Xcal_j$ be the product spaces of any measurable spaces and
$$ \Pr_V(X_V|X_J)$$ 
a Markov kernel (i.e.\ a family of distributions on $\Xcal_V$ measurably\footnote{We require that for every measurable $F \ins \Xcal_V$ the map $\Xcal_J \to [0,1]$ given by $x_J \mapsto \Pr_V(X_V \in F|X_J=x_J)$ is measurable.} parametrized by $\Xcal_J$).
For subsets $A,B,C \ins V \dot \cup J$ we write:
$$ X_A \Indep_{\Pr_V(X_V|X_J),\bullet} X_B \given X_C$$
if and only if for \emph{every joint} distribution $\Pr_J$ on $\Xcal_J$ we have:
$$X_A \Indep_{\Pr_{V \cup J}} X_B \given X_C,$$
which means that for all measurable $F \ins \Xcal_A$ we have:
$$ \Pr_{V \cup J}(X_A\in F|X_B,X_C) = \Pr_{V \cup J}(X_A \in F|X_C) \quad \Pr_{V \cup J}\text{-a.s.,} $$
where $\Pr_{V \cup J}(X_{V\cup J}):=\Pr_V(X_V|X_J) \otimes \Pr_J(X_J)$, the distribution given by $X_J \sim \Pr_J$ and then $X_V \sim \Pr_V(\_|X_J)$. 
\end{Def}

\begin{Lem}
\label{ci-input-confounded-lem}
Let the situation be like in \ref{def-input-confounded-cond-ind} and assume all spaces $\Xcal_v$, $v \in V$, to be standard measurable spaces.
Let $A,B,C$ be pairwise disjoint, $A \cap J = \emptyset$ and $J \ins B \cup C$.
Then every statement implies the one below:
\begin{enumerate}
	\item There is a version of $\Pr_V(X_A|X_B,X_C)$ such that for all $x_B,x_B' \in \Xcal_B$, $x_C \in \Xcal_{C}$:
	\begin{align*}
&\;\Pr_V(X_A|X_B=x_B,X_C=x_C) \\
=&\; \Pr_V(X_A|X_B=x_B',X_C=x_C). \qedhere
\end{align*}
	\item $\displaystyle X_A \Indep_{\Pr_V(X_V|X_J),\bullet} X_B \given X_C$. 
	\item $\displaystyle X_A \Indep_{\Pr_V(X_V|X_J)} X_B \given X_C$ (using definition \ref{def-cond-ind}).
	\item $\displaystyle X_A \Indep_{\Pr_V(X_V|X_J)\otimes \delta_{x_J}(X_J)} X_B \given X_C$ for every $x_J \in \Xcal_J$.
\end{enumerate}
If there is a Markov kernel $\Pr(X_A|X_C)$ that is a version of $\Pr_{V \cup J}(X_A|X_C)$ for every Dirac delta distribution $\Pr_J=\delta_{x_J}$ 
(e.g.\ if $J \ins C$) then the last point also implies the first.
\begin{proof} 1. $\implies$ 2.: Functional dependence only on $x_C$.\\
2. $\implies$ 3. $\implies$ 4.: Every product distribution is a joint distribution and every Dirac delta distribution is a product distribution.\\
1. $\Longleftarrow$ 4.: Let $N \ins \Xcal_{B \cup C}$ be the measurable set on which 
the Markov kernels $\Pr_V(X_A|X_B,X_C)$ and $\Pr(X_A|X_C)$ (considered as functions of $(x_B,x_C)$) differ. For every $x_J \in \Xcal_J$ we have by assumption:
$$X_A \Indep_{\Pr_V(X_V|X_J)\otimes\delta_{x_J}(X_J)} X_B \given X_C.$$
This shows that:
$$ \Pr_V(X_A|X_B=x_B,X_C=x_C) = \Pr(X_A|X_C=x_C) $$
for $(x_B,x_C)$ outside of a $\Pr_V(X_{(B \cup C) \sm J}|X_J=x_J)$-zero set, for which we can take the section $N_{x_J}$ of $N$.
This implies that $N$ is a $\Pr_V(X_{(B \cup C) \sm J}|X_J)$-zero set.
So $\Pr(X_A|X_C)$ is a version of $\Pr_V(X_A|X_B,X_C)$ and  satisfies 1..
\end{proof}
\end{Lem}

\begin{Rem}
\begin{enumerate}
\item The existence of the Markov kernel $\Pr(X_A|X_C)$  under the assumption 4.\ in lemma \ref{ci-input-confounded-lem}
always/only holds up to measurability questions, because for every fixed $\Pr_J$ the regular conditional probability distribution $\Pr_{V\cup J}(X_A|X_B,X_C)$ always exists in standard measurable spaces and agrees with $\Pr_{V\cup J}(X_A|X_C)$ (by the assumption 4.). 
The existence of the Markov kernel $\Pr(X_A|X_C)$ follows for standard measurable spaces $\Xcal_v$, $v \in V$, if either:
\begin{enumerate}
  \item $J \ins C$ and assumption 4. holds, or:
	\item $\Xcal_J$ is discrete and assumption 2. holds, or:
	\item $\Pr_V(X_V|X_J)$ comes as $\Pr_U(X_V|X_J)$ from an \ioSCM\ and assumptions 2.-4. even hold in form of the corresponding $\sigma$-separation statement in the induced DMG $G$. 
\end{enumerate}   
We plan in future work to address all these subtleties in more detail.
\item Lemma \ref{ci-input-confounded-lem} shows that definition \ref{def-input-confounded-cond-ind} (and also already definition \ref{def-cond-ind})  generalizes the one from \cite{RERS17s} (when applied symmetrized). The clear correspondence/generalization is that for any (not necessarily disjoint) $A,B,C \ins V \cup J$:
\begin{align*} 
&&\quad\;X_A \Indep_{\text{\cite{RERS17s}}} X_B \given X_C \\
&:\iff &\quad\;X_A \Indep_{\Pr_V(X_V|X_J),\bullet} X_{B \cup J} \given X_C \\
&&\lor\quad X_B \Indep_{\Pr_V(X_V|X_J),\bullet} X_{A \cup J} \given X_C. 
\end{align*}
\item Thm.\ 4.4 in  \cite{CD17s} shows that definitions \ref{def-cond-ind}, \ref{def-input-confounded-cond-ind} also generalize the one from \cite{CD17s} in the same sense.
\item In contrast with \cite{Daw79, CD17s, RERS17s}, definition \ref{def-input-confounded-cond-ind} can accommodate any variable from $V$ or $J$ at any position of the conditional independence statement.
\item Also note that $\Indep_{\Pr_V(X_V|X_J),\bullet}$ is well-defined for any measurable spaces and is not restricted to discrete variables or distributions/Markov kernels that come with densities. 
\item Furthermore, $\Indep_{\Pr_V(X_V|X_J),\bullet}$ satisfies the \emph{separoid axioms} (see \cite{Daw79,PeaPaz87,GeiVerPea90} or see rules 1-5 in Lem.\ \ref{GrAxPr} for $\Indep_{\Pr_V(X_V|X_J),\bullet}$).
Indeed, every single $\Indep_{\Pr_{V \cup J}}$ satisfies the separoid axioms (see \cite{Daw79,CD17s}) and an arbitrary intersection of separoids is again a separoid (see \cite{Daw01}): %
$$\left\langle\Indep_{\Pr_V(X_V|X_J),\bullet}\right\rangle = \bigcap_{\Pr_J}\left\langle \Indep_{\Pr_{V \cup J}}\right\rangle.$$
\end{enumerate}
\end{Rem}

\subsection{INPUT CONFOUNDED GLOBAL MARKOV PROPERTY}

We can also prove a global Markov property for the input confounded version of conditional independence. For this we need to modify the graphical structures a bit and introduce a few more notations. Note that all spaces are assumed to measurable (but not necessarily standard).

\begin{Def}[Input confounded \ioSCM]
Let $M=(G^+,\Xcal,\Pr_U,g)$ be an \ioSCM\ with graph $G^+=(V \dot \cup U\dot \cup J, E^+)$.
The corresponding \emph{input confounded \ioSCM} $M_\bullet$ is then constructed from $M$ by the following changes:
\begin{enumerate}
	\item $V_\bullet := V \cup J$ and $U_\bullet := U$,
	\item $J_\bullet := \{ \bullet \}$ with a new node $\bullet$ with space $\Xcal_\bullet := \Xcal_J$,
	\item $E_\bullet^+ := E^+ \cup \{ \bullet \tuh j\,|\, j \in J\}$,
	\item add $g_{\{j\}}$, the canonical projection from $\Xcal_\bullet$ onto $\Xcal_j$, to $g$ for $j \in J$.
\end{enumerate}
With this setting $M_\bullet$ is a well-defined \ioSCM.\\
Furthermore, let $G_\bullet$ be the \emph{input confounded induced DMG}, i.e.\ the induced DMG of $G^+_\bullet$ 
 where $\bullet$ is marginalized out. 
In other words, $G_\bullet$ arises from the induced DMG $G$ of $G^+$ by just adding 
$j_1 \huh j_2$ for all $j_1,j_2 \in J$, $j_1 \neq j_2$, to $G$.
\end{Def}

\begin{Thm}[Input confounded directed global Markov property]
\label{input-confounde-mSCM-gdGMP-thm} 
Let $M$ be an \ioSCM\ with input confounded induced DMG $G_\bullet$.
Then for all subsets $A,B,C \ins V \cup J$ we have the implication:
\[ A \Indep^\sigma_{G_\bullet} B \given C \;\implies\; X_A \Indep_{\Pr_U(X_V|\doit(X_J)),\bullet} X_B \given X_C. \]
In words, if $A$ and $B$ are $\sigma$-separated by $C$ in $G_\bullet$ then the corresponding variables $X_A$ and $X_B$ are conditionally independent given $X_C$ for any distribution $\Pr_U(X_V|\doit(X_J))\otimes \Pr_J(X_J)$ for any \emph{joint} distribution $\Pr_J$ on $\Xcal_J$.
\begin{proof}
This directly follows from the $\sigma$-separation criterion/global Markov property \ref{mSCM-gdGMP-thm} applied to the input confounded \ioSCM\ $M_\bullet$ and $G^+_\bullet$, or, alternatively, again from the mSCM-version proven in \cite{FM18,FM17} for each fixed joint distribution $\Pr_J$ on $\Xcal_J=\Xcal_\bullet$. Note that $G_\bullet$ is a marginalization of $G^+_\bullet$ and $\sigma$-separation is stable under marginalization.
\end{proof}
\end{Thm}

\section{THE EXTENDED IOSCM - PROOFS}

\begin{Prp}
\label{caus-calc-eqn-proof}
Let $M=(G^+,\Xcal,\Pr_U,g)$ be an \ioSCM\ with $G^+=(V \dot \cup U \dot \cup J, E^+)$ and $\hat M$ the extended \ioSCM.
Let $A,B,C \ins V$ 
be pairwise disjoint set of nodes and $x_{C\cup J} \in \Xcal_{C \cup J}$. Then we have the equations:
\begin{align*} 
 &\; \Pr_U(X_A|X_B, \doit(X_{C\cup J}=x_{C \cup J})) \\
=&\; \Pr_U (X_A | X_B, I_C=x_C,X_J=x_J) \\
 =&\; \Pr_U (X_A | X_B, I_C=x_C, X_C=x_C,X_J=x_J).
\end{align*}
\begin{proof}%
Consider the first equality. 
For any subset $D \ins V$ the variable $X_D^{\doit(X_{C \cup J}=x_{C \cup J})}$ was recursively defined in $M_{\doit(C)}$ via $g$ using $G^+_{\doit(C)}$, 
whereas the variable 
$X_D^{\doit((I_C,I_{V\sm C},X_J)=(x_C,\obs_{V\sm C},x_J))}$ 
was recursively defined in $\hat M$ via the same $g$ but using $I(x_C,\obs_{V \sm C})$ and $G^+_{\doit(I(x_C,\obs_{V \sm C}))}$. 
Since $x_C \in \Xcal_C$ we have that $I(x_C,\obs_{V \sm C})=C$ and thus 
$G^+_{\doit(I(x_C,\obs_{V \sm C}))}=G^+_{\doit(C)}$.  It directly follows that:
\begin{align*}
 X_D^{\doit(X_{C \cup J}=x_{C \cup J})} &= X_D^{\doit((I_C,I_{V\sm C},X_J)=(x_C,\obs_{V\sm C},x_J))}.
\end{align*}
This shows the equality of top and middle line.
For the equality between the middle and bottom line note that:
$$I_C=x_C   \stackrel{x_C\in \Xcal_C}{\implies} X_C=x_C. \eqno\qedhere$$
\end{proof}
\end{Prp}

\section{THE THREE MAIN RULES OF CAUSAL CALCULUS - PROOFS}

\begin{Thm}[The three main rules of causal calculus]
\label{three-rules-proofs}
Let $M$ be an \ioSCM\ with set of observed nodes $V$ and intervention nodes $J$ and induced DMG $G$.
Let $X,Y,Z \ins V$ and $J \ins W \ins V \cup J$ be subsets. 
\begin{enumerate}
\item Insertion/deletion of observation: 
\item[] If $\qquad\displaystyle Y \Indep_{ G }^\sigma X |Z, \doit(W)\qquad$ then:
 $$\Pr(Y|X, Z, \doit(W)) = \Pr(Y| Z, \doit(W)).$$
\item Action/observation exchange:
\item[] If $\qquad\displaystyle Y \Indep_{G}^\sigma I_X | X, Z, \doit(W)\qquad$ then:
$$\Pr(Y| \doit(X), Z,  \doit(W)) = \Pr(Y|X, Z,  \doit(W)).$$
\item Insertion/deletion of actions:
\item[] If $\qquad\displaystyle Y \Indep_{G}^\sigma I_X | Z,  \doit(W)\qquad$ then:
$$\Pr(Y| \doit(X), Z,  \doit(W)) = \Pr(Y| Z, \doit(W)).$$
\end{enumerate}
\begin{proof}
\begin{enumerate}
\item Thm.\ \ref{mSCM-gdGMP-thm} applied to $G_{\doit(W)}$ gives:
$$  Y \Indep_{ G }^\sigma X |Z, \doit(W) \;\stackrel{\ref{mSCM-gdGMP-thm}}{\implies}\; Y \Indep_\Pr X |Z, \doit(W). $$
The latter directly gives the claim:
$$\Pr(Y|X, Z, \doit(W)) = \Pr(Y| Z, \doit(W)).$$
\item The $\sigma$-separation criterion \ref{mSCM-gdGMP-thm} w.r.t.\ to $\hat G_{\doit(W)}$ gives:
$$  Y \Indep_{G}^\sigma I_X | X, Z, \doit(W) \,\stackrel{\ref{mSCM-gdGMP-thm}}{\implies}\, Y \Indep_\Pr I_X | X, Z, \doit(W). $$
Together with Prp. \ref{caus-calc-eqn} (applied to $M_{\doit(W)}$) we have:
\begin{eqnarray*}
&& \Pr(Y| \doit(X), Z,  \doit(W)) \\
&\stackrel{\ref{caus-calc-eqn}}{=}& \Pr(Y| I_X, X, Z,  \doit(W)) \\
&\stackrel{Y \Indep I_X | X, Z, \doit(W)}{=}& \Pr(Y| X, Z,  \doit(W)).
\end{eqnarray*}
\item As before we have:
$$ Y \Indep_{G}^\sigma I_X |  Z, \doit(W) \;\stackrel{\ref{mSCM-gdGMP-thm}}{\implies}\; Y \Indep_\Pr I_X | Z, \doit(W).$$
And again: $\qquad\Pr(Y| \doit(X), Z,  \doit(W))$
\begin{eqnarray*}
&\stackrel{\ref{caus-calc-eqn}}{=}& \Pr(Y| I_X, Z,  \doit(W)) \\
&\stackrel{Y \Indep I_X | Z, \doit(W)}{=}& \Pr(Y| Z,  \doit(W)). \qquad\qquad \qedhere
\end{eqnarray*}
\end{enumerate}
\end{proof}

\end{Thm}

\section{ADJUSTMENT CRITERIA}
\subsection{PROOFS}

\begin{Thm}[General adjustment criterion and formula]
\label{adjustment-proofs}
Let the setting be like in \ref{selection-backdoor-not}.
Assume that data was collected under selection bias, $\Pr(V|S=s,\doit(W))$ (or under $\Pr(V|\doit(W))$ and $S=\emptyset$), and there are unbiased samples from $\Pr(Z|C,\doit(W))$.
Further assume that the variables satisfy:
\begin{enumerate}
\item $\displaystyle (Z_0,L) \Indep^\sigma_{G} I_X \given C,\doit(W)$, and
\item $\displaystyle Y \Indep^\sigma_{G} (I_X, Z_+) \given C, X, Z_0,L,\doit(W)$, and
\item $\displaystyle Y \Indep^\sigma_{G} S \given C,X, Z,\doit(W)$, and 
\item $\displaystyle L \Indep^\sigma_{G} X \given C,Z,\doit(W)$.
\end{enumerate}
Then one can estimate the conditional causal effect $\Pr(Y|C,\doit(X),\doit(W))$ via the \emph{adjustment formula}:
\begin{equation*}\begin{split}
  & \Pr(Y|C,\doit(X),\doit(W))  \\
&= \int \Pr(Y | X,Z, C, S=s,\doit(W))\, d\Pr(Z|C,\doit(W)).
\end{split}\end{equation*}
\begin{proof}
Since $C$, $\doit(W)$ occur everywhere as a conditioning set, we will suppress $C$, $\doit(W)$ in the following everywhere.
Then note that the $\sigma$-separation criterion \ref{mSCM-gdGMP-thm} implies the corresponding conditional independencies in the following when indicated.
The adjustment formula then derives from the following computations:
\begin{eqnarray*}
&& \Pr(Y|\doit(X)) \\
&=& \int \Pr(Y| Z_0,L,\doit(X)) \,\\&&\qquad\qquad d\Pr(Z_0,L|\doit(X)) \\
&\stackrel{\ref{caus-calc-eqn}}{=} & \int \Pr(Y|I_X, X, Z_0,L) \,  d\Pr(Z_0,L|I_X) \\
&\stackrel[(Z_0,L) \Indep I_X]{Y\Indep I_X | X,Z_0,L;}{=} & \int \Pr(Y| X, Z_0,L) \, d\Pr(Z_0,L) \\
&\stackrel{\int d\Pr(Z_+|Z_0,L) = 1}{=}& \int \int \Pr(Y| X, Z_0,L) \, \\&&\qquad \qquad d\Pr(Z_+|Z_0,L) \, d\Pr(Z_0,L) \\
&\stackrel{ Y\Indep Z_+ | X,Z_0,L }{=}& \int \Pr(Y| X,Z_0,Z_+,L) \, d\Pr(Z_+,Z_0,L) \\
&\stackrel{Z = Z_+ \cup Z_0}{=}& \int \Pr(Y| X,Z,L) \, d\Pr(Z,L) \\
&\stackrel{}{=}& \int\int \Pr(Y| X,Z,L) \,  d\Pr(L|Z) \,d\Pr(Z) \\
&\stackrel{L\Indep X| Z}{=}& \int\int \Pr(Y|L,X,Z) \,\\&&\qquad\qquad  d\Pr(L|X,Z) \,d\Pr(Z) \\ %
\end{eqnarray*}
\begin{eqnarray*}
&\stackrel{}{=}& \int \Pr(Y|X,Z) \,d\Pr(Z) \\
&\stackrel{ Y\Indep S | X,Z }{=}& \int \Pr(Y| X,Z,S) \, d\Pr(Z).\qedhere
\end{eqnarray*}
\end{proof}
\end{Thm}

\subsection{SPECIAL CASES}

\begin{Cor}
\label{backdoor-cor-supp}
Let the notations be like in \ref{selection-backdoor-not} and \ref{adjustment} and $W=J=\emptyset$.
We have the following special cases, which in the \emph{acyclic} case will reduce to the ones given by the indicated references:
\begin{enumerate}
\item General selection-backdoor (see \cite{CB17s}): $C=\emptyset$,
 and 
\begin{enumerate}
\item $\displaystyle (Z_0,L) \Indep^\sigma_{G} I_X$, and
\item $\displaystyle Y \Indep^\sigma_{G} (I_X, Z_+) \given X, Z_0,L$, and
\item $\displaystyle Y \Indep^\sigma_{G} S \given X, Z$, and 
\item $\displaystyle L \Indep^\sigma_{G} X \given Z$, implies:
\end{enumerate}
 $$\displaystyle \Pr(Y|\doit(X)) = \int \Pr(Y|X,Z,S=s)\, d\Pr(Z).$$

\item Selection-backdoor (see \cite{BTP14s}): $C=L=\emptyset$, and 
 \begin{enumerate}
\item $\displaystyle Z_0 \Indep^\sigma_{G} I_X$, and
\item $\displaystyle Y \Indep^\sigma_{G} (I_X, Z_+, S) \given X, Z_0$ implies:
\end{enumerate}
 $$\displaystyle \Pr(Y|\doit(X)) = \int \Pr(Y|X,Z,S=s)\, d\Pr(Z).$$

\item Extended backdoor\footnote{In the acyclic case it was shown in \cite{SWR10s} that when $L$ is allowed to represent latent variables in a graph $G'$ that marginalizes to $G$ then this criterion actually characterizes all adjustment sets for $G$ and $\Pr(Y|\doit(X))$.}  (see \cite{PearlPaz13,SWR10s}): $C=S=\emptyset$, %
\begin{enumerate}
\item $\displaystyle (Z_0,L) \Indep^\sigma_{G} I_X$, and
\item $\displaystyle Y \Indep^\sigma_{G} (I_X, Z_+) \given X, Z_0,L$, and
\item $\displaystyle L \Indep^\sigma_{G} X \given Z$, implies:
\end{enumerate}
 $$\displaystyle \Pr(Y|\doit(X)) = \int \Pr(Y|X,Z)\, d\Pr(Z).$$

\item Backdoor (see \cite{Pearl93as,Pearl93bs,Pearl09}):  $C=S=L=Z_+=\emptyset$,%
\begin{enumerate}
\item $\displaystyle Z \Indep^\sigma_{G} I_X$, and %
\item $\displaystyle Y \Indep^\sigma_{G} I_X \given X, Z$,  implies:
\end{enumerate}
$$\displaystyle \Pr(Y|\doit(X)) = \int \Pr(Y|X,Z)\, d\Pr(Z).$$
\end{enumerate}
\end{Cor}

\subsection{MORE ON ADJUSTMENT CRITERIA}

The following generalizes the adjustment criterion of type I in \cite{CB17s}.

\begin{Thm}[General adjustment without external data]
\label{another-adjustment-I}
Let the setting be like in \ref{selection-backdoor-not}.
Assume that data was collected under selection bias, $\Pr(V|S=s)$.
Further assume that the variables satisfy:
\begin{enumerate}
\item $\displaystyle Y \Indep^\sigma_{G} S \given \doit(X)$,
\item $\displaystyle Z_0 \Indep^\sigma_{G} I_X \given S$,
\item $\displaystyle Y\Indep^\sigma_{G} Z_+ \given Z_0,S,\doit(X)$,
\item $\displaystyle Y\Indep^\sigma_{G} I_X \given X, Z,S$.
\end{enumerate}
Then one can estimate the causal effect $\Pr(Y|\doit(X))$ via the following \emph{adjustment formula} from the biased data:
$$\Pr(Y|\doit(X)) = \int \Pr(Y | X,Z, S=s)\, d\Pr(Z|S=s).$$
\begin{proof}
First note that the $\sigma$-separation criterion Theorem \ref{mSCM-gdGMP-thm} implies the corresponding conditional independencies in the following when indicated. We implicitly make use of Proposition \ref{caus-calc-eqn} when needed.
The adjustment formula then derives from the following computations:
\begin{eqnarray*}
&& \Pr(Y|\doit(X)) \\
&\stackrel{Y \Indep S \given \doit(X)}{=}& \Pr(Y|S,\doit(X)) \\
&\stackrel{\text{chain rule}}{=}& \int \Pr(Y| Z_0,S,\doit(X))  \\& &\,d\Pr(Z_0|S,\doit(X)) \\
&\stackrel[\ref{caus-calc-eqn}]{Z_0 \Indep I_X \given S}{=}& \int \Pr(Y| Z_0,S,\doit(X))  \,d\Pr(Z_0|S) \\
&\stackrel{\int d\Pr(Z_+|Z_0,S) = 1}{=}& \int \Pr(Y| Z_0,S,\doit(X)) \\& &\, d\Pr(Z_+,Z_0|S) \\
&\stackrel{Y\Indep Z_+ \given Z_0,S,\doit(X)}{=}& \int \Pr(Y| Z_+,Z_0,S,\doit(X)) \\& &\, d\Pr(Z_+,Z_0|S) \\
&\stackrel{Z = Z_+ \cup Z_0}{=}& \int \Pr(Y| Z,S,\doit(X)) \, d\Pr(Z|S) \\
&\stackrel[\ref{caus-calc-eqn}]{Y \Indep I_X \given X, Z, S}{=}& \int \Pr(Y| Z,S,X) \, d\Pr(Z|S).
\end{eqnarray*}
\end{proof}
\end{Thm}

The following theorem generalizes the adjustment criterion of type III in \cite{CTB18s}. 
For this we have to introduce even more adjustment sets: $Z_0^A,Z_0^B,Z_1^A,Z_1^B,Z_2,Z_3$ and $L_0,L_1$.
We write $Z_0 = (Z_0^A,Z_0^B)$, $Z_{\le1}^A = (Z_0^A,Z_1^A)$, etc..

\begin{Thm}[General adjustment with partial external data]
\label{another-adjustment-III}
Assume that data was collected under selection bias, $\Pr(V|S=s)$, but we have unbiased data from $\Pr(Z^B_{\le1})$.
Further assume that the variables satisfy:
\begin{enumerate}
\item $\displaystyle (L_0,Z_0) \Indep I_X$,
\item $\displaystyle Y \Indep Z_1 \given L_0,Z_0,\doit(X)$,
\item $\displaystyle Z_{\le 1}^A \Indep S \given Z_{\le 1}^B$,
\item $\displaystyle L_0 \Indep I_X \given Z_{\le 1}$,
\item $\displaystyle Y \Indep S \given Z_{\le1},\doit(X)$,
\item $\displaystyle (L_1,Z_2) \Indep I_X \given S,Z_{\le1}$,
\item $\displaystyle Y \Indep Z_3 \given L_1,S,Z_{\le2},\doit(X)$,
\item $\displaystyle L_1 \Indep I_X \given S,Z$,
\item $\displaystyle Y \Indep I_X \given X,S,Z$.
\end{enumerate}
Then we have the adjustment formula: $\;\Pr(Y|\doit(X)) = $
$$\int\int \Pr(Y|S=s,Z,X) \,d\Pr(Z \sm Z_{\le1}^B|S=s,Z_{\le 1}^B) \, d\Pr(Z_{\le1}^B).$$
Note that this formula does not depend on $L_0$ and $L_1$. So $L_0$ and $L_1$ can be chosen in a graph $G'$ that marginalizes to $G$.
\begin{proof}
\begin{eqnarray*}
&& \Pr(Y|\doit(X)) \\
&\stackrel{\text{chain rule}}{=}& \int \Pr(Y| L_0,Z_0,\doit(X)) \\&&\qquad d\Pr(L_0,Z_0|\doit(X)) \\
&\stackrel[\ref{caus-calc-eqn}]{(L_0,Z_0) \Indep I_X}{=}& \int \Pr(Y| L_0,Z_0,\doit(X))  \, \\&&\,d\Pr(L_0,Z_0) \\
&\stackrel[Z_{\le1}=Z_0 \cup Z_1]{\int d\Pr(Z_1|L_0,Z_0) = 1}{=}& \int \Pr(Y| L_0,Z_0,\doit(X)) \, \\&&\, d\Pr(L_0,Z_{\le 1}) \\
&\stackrel{Y \Indep Z_1 \given L_0,Z_0,\doit(X)}{=}& \int \Pr(Y| L_0,Z_{\le1},\doit(X)) \,  \\&&\,d\Pr(L_0,Z_{\le 1}) \\
&\stackrel[Z_{\le1}=Z^A_{\le1} \cup Z^B_{\le 1}]{\text{chain rule}}{=}& \int \Pr(Y| L_0,Z_{\le1},\doit(X)) \\
&&\, d\Pr(L_0|Z_{\le 1})\, d\Pr(Z_{\le1}^A|Z_{\le 1}^B) \, \\&&\, d\Pr(Z_{\le1}^B) \\
&\stackrel{Z_{\le 1}^A \Indep S \given Z_{\le 1}^B}{=}& \int \Pr(Y| L_0,Z_{\le1},\doit(X)) \\
&&\, d\Pr(L_0|Z_{\le 1})\,  d\Pr(Z_{\le1}^A|S,Z_{\le 1}^B) \, \\&&\, d\Pr(Z_{\le1}^B) \\
&\stackrel[\ref{caus-calc-eqn}]{L_0 \Indep I_X \given Z_{\le 1}}{=}& \int \Pr(Y| L_0,Z_{\le1},\doit(X)) \\
&&\, d\Pr(L_0|Z_{\le 1},\doit(X)) \\&&\, d\Pr(Z_{\le1}^A|S,Z_{\le 1}^B) \, d\Pr(Z_{\le1}^B) \\
&\stackrel{\text{chain rule}}{=}& \int \Pr(Y| Z_{\le1},\doit(X)) \\
&&\qquad d\Pr(Z_{\le1}^A|S,Z_{\le 1}^B) \, d\Pr(Z_{\le1}^B) \\
\end{eqnarray*}
\begin{eqnarray*} 
&\stackrel{Y \Indep S \given Z_{\le1},\doit(X) }{=}& \int \Pr(Y|S,Z_{\le1},\doit(X)) \\
&&\, d\Pr(Z_{\le1}^A|S,Z_{\le 1}^B) \, d\Pr(Z_{\le1}^B) \\ 
&\stackrel{\text{chain rule}}{=}& \int \Pr(Y|L_1,Z_2,S,Z_{\le1},\doit(X)) \\
&&\, d\Pr(L_1,Z_2|S,Z_{\le1},\doit(X))  \\
&&\, d\Pr(Z_{\le1}^A|S,Z_{\le 1}^B) \, d\Pr(Z_{\le1}^B) \\
&\stackrel{Z_{\le2}=Z_{\le1}\cup Z_2}{=}& \int \Pr(Y|L_1,S,Z_{\le2},\doit(X)) \\
&&\, d\Pr(L_1,Z_2|S,Z_{\le1},\doit(X))  \\
&&\, d\Pr(Z_{\le1}^A|S,Z_{\le 1}^B) \, d\Pr(Z_{\le1}^B) \\
&\stackrel[\ref{caus-calc-eqn}]{(L_1,Z_2) \Indep I_X \given S,Z_{\le1}}{=} & \int \Pr(Y|L_1,S,Z_{\le2},\doit(X)) \\
&&\, d\Pr(L_1,Z_2|S,Z_{\le1})  \\
&&\, d\Pr(Z_{\le1}^A|S,Z_{\le 1}^B) \, d\Pr(Z_{\le1}^B) \\
&\stackrel[\int \Pr(Z_3|L_1,S,Z_{\le2})=1]{Y \Indep Z_3 \given L_1,S,Z_{\le2},\doit(X) }{=} & \int \Pr(Y|L_1,S,Z_{\le2},Z_3,\doit(X)) \\
&&\, d\Pr(L_1,Z_2,Z_3|S,Z_{\le1})  \\
&&\, d\Pr(Z_{\le1}^A|S,Z_{\le 1}^B) \, d\Pr(Z_{\le1}^B) \\
&\stackrel[Z=Z_{\le 2}\cup Z_3]{\text{chain rule}}{=} & \int \Pr(Y|L_1,S,Z,\doit(X)) \\
&&\, d\Pr(L_1|S,Z)  \\
&&\, d\Pr(Z \sm Z_{\le1}^B|S,Z_{\le 1}^B) \, d\Pr(Z_{\le1}^B) \\
&\stackrel[\ref{caus-calc-eqn}]{L_1 \Indep I_X \given S,Z }{=} & \int \Pr(Y|L_1,S,Z,\doit(X)) \\
&&\, d\Pr(L_1|S,Z,\doit(X))  \\
&&\, d\Pr(Z \sm Z_{\le1}^B|S,Z_{\le 1}^B) \, d\Pr(Z_{\le1}^B) \\
\end{eqnarray*}
\begin{eqnarray*}
&\stackrel{\text{chain rule}}{=} & \int \Pr(Y|S,Z,\doit(X)) \\
&&\, d\Pr(Z \sm Z_{\le1}^B|S,Z_{\le 1}^B) \, d\Pr(Z_{\le1}^B) \\
&\stackrel{Y \Indep I_X \given X,S,Z}{=} & \int \Pr(Y|S,Z,X) \\
&&\, d\Pr(Z \sm Z_{\le1}^B|S,Z_{\le 1}^B) \, d\Pr(Z_{\le1}^B).
\end{eqnarray*}
\end{proof}
\end{Thm}

\section{IDENTIFYING CAUSAL EFFECTS}

\begin{Rem}[More remarks about the ID-algorithm] $ $
\label{id-rem-supp}
\begin{enumerate}
  \item The extended version of the ID algorithm is equivalent to applying the ID algorithm to the \emph{acyclification} $G^{+,\acy}$ of $G^+$, which here is meant to be the conditional ADMG that arises by adding edges $v \tuh w'$ if $v \notin \Sc^G(w) \ni w'$ and $ v\tuh w \in G^+$, and erasing all edges inside $\Sc^G(w)$, $w \in V$ (see \cite{FM17}). 
	\item A consolidated district in $G$ then is the same as a district in $G^\acy$.
	\item Every apt-order of $G$ is a topological order of $G^\acy$.
	\item So identifiability in $G^\acy$ implies identifiability in $G$.
	\item This leads to the rule of thumb that causal effects where both cause and effect nodes are inside one strongly connected component of $G$ are not identifiable from observational data alone, and, that the causal effects of sets of nodes between strongly connected components follow rules similar to the acyclic case.
		\item Similarly, the corner cases for the identification of \emph{conditional} causal effects $\Pr(Y|R,\doit(W))$ in $G$ that are not covered by the identification of $\Pr(Y,R|\doit(W))$ in $G$ follow from the (acyclic) conditional ID-algorithm from \cite{Tian04s} applied to $G^\acy$ and then translated back to $G$ by the above correspondences.
\end{enumerate}

\end{Rem}

\begin{Lem}
\label{new-central-lemma}
Let $M=(G^+,\Xcal,\Pr_U,g)$ be an \ioSCM\ with $G^+=(V \dot \cup U \dot \cup J, E^+)$ and 
$<$ an apt-order for $G^+$ and $G$ its induced DMG (with nodes $V \dot \cup J$). 
Let $S \ins V$ be a strongly connected component of $G$ and $D \ins V$ be any union of consolidated districts in $G$ with $S \ins D$ 
(e.g.\ $D=\CDist^G(S)$) and $P:=\Pa^G(D)\sm D$.
Then we have the equality (indices for emphasis):
\begin{align*}
& \;\Pr_M(S|\Pred_<^G(S)\cap V,\doit(J)) \\&= 
\Pr_{M_{[D]}}(S|\Pred_<^{G_{[D]}}(S) \cap D,\doit(P)).
\end{align*}
\begin{proof}
First note that since $D$ is a union of strongly connected components and all other variables in $G_{[D]}$ have no parents the total order $<$ is also an apt-order for $G_{[D]}$. It follows that we have the equality of sets of nodes:
\begin{align*}
\Pred^{G_{[D]}}_<(S) \cap D&= \Pred^G_<(S) \cap D &=: D_<.
\end{align*}
Now we introduce the following further abbreviations:
\begin{align*}
D_> & := D \sm ( S \cup D_< ), \\
P_< &:= \Pred^G_<(S) \cap (P \cap V), \\
P_> &:= (P \cap V) \sm \Pred^G_<(S), \\
P_J &:= P \cap J, \\
J_< &:= \Pred^G_<(S) \cap J, \\
J_> &:= J \sm \Pred^G_<(S), \\
R_< &:= \Pred^G_<(S) \cap V \sm ( D  \cup P),\\
R_> &:= V \sm (D \cup P \cup \Pred^G_<(S)).
\end{align*}
Then we get the relations between the sets of nodes:
\begin{align*}
 V &= R_< \,\dot\cup\, D \,\dot\cup\, R_> \,\dot\cup\, P_< \,\dot\cup\, P_>\\
 D &= D_< \,\dot\cup\, S \,\dot\cup\, D_>, \\
 P &= P_< \,\dot\cup\, P_> \,\dot\cup\, P_J, \\
\Pred^G_<(S) \cap V &= D_< \,\dot\cup\, R_< \,\dot\cup\, P_<,\\
J & = J_< \,\dot\cup\, J_>.
\end{align*}
Since $\Pred^G_\le(S)$ is ancestral in $G$ and $\Pred_\le^{G_{[D]}}(S)$ is ancestral in $G_{[D]}$, resp., we can by remark \ref{sub-iSCM-remark}  arbitrarily intervene on all variables outside of these sets without changing the distributions $\Pr_M(S|\Pred_<^G(S) \cap V,\doit(J))$ and $\Pr_{M_{[D]}}(S|\Pred_<^{G_{[D]}}(S) \cap D,\doit(P))$, resp..
With these remarks and our new notations we have the equalities:
\begin{align*}
&\; \Pr_M(S|\Pred_<^G(S) \cap V,\doit(J)) \\
&= \Pr_M(S|D_<,R_<,P_<,\doit(J)) \\
 &\stackrel{\ref{sub-iSCM-remark}}{=} \Pr_M(S|D_<,R_<,P_<,\doit(J,R_>, P_>, D_>));
\end{align*}
and:
\begin{align*}
&\; \Pr_{M_{[D]}}(S|\Pred_<^{G_{[D]}}(S)\cap D,\doit(P)) \\
&= \Pr_{M_{[D]}}(S|D_<,\doit(P_<,P_>,P_J))\\
 &\stackrel{\ref{sub-iSCM-remark}}{=} \Pr_{M_{[D]}}(S|D_<,\doit(P_<,P_>,P_J,D_>)) \\
&\stackrel{\ref{sub-iSCM-remark}}{=} \Pr_{M}(S|D_<,\doit(P_<,P_>,J,D_>,R_<,R_>)).
\end{align*}
So the equality between those expressions and thus the claim follows by the 2nd rule of causal calculus in Theorem \ref{three-rules} with the $\sigma$-separation statement:
\[ S \Indep_G^\sigma I_{R_<,P_<} \given D_<,R_<,P_<,  \doit(J,R_>, P_>, D_>). \]
To prove the latter note that the intervention $\doit(R_>, P_>, D_>)$ allows us to restrict to the ancestral subgraph $\Pred^G_\le(S) \cup J$.
Now let $\pi$ be a path from an indicator variable from $I_{R_<,P_<}$ to $S$ (in $\Pred^G_\le(S) \cup J$).
Then the path can only be of the form:
\[ v_i \cdots v_p \tuh v_d \cdots v_s, \]
with $v_i \in I_{R_<,P_<}$, $v_p \in P_<$, $v_d \in D$, $v_s \in S$, as there cannot be any bidirected edge or directed edge in the other direction between $R_< \cup P_<$ and $D$ by the definition of consolidated districts and $P=\Pa^G(D) \sm D$. Since we condition on $P_<$ the path $\pi$ is $\sigma$-blocked.
\end{proof}
\end{Lem}

\begin{Rem}
Another way to deal with the problem that consolidated districts are not topologically ordered in the extended ID-algorithm (see Algorithm~\ref{ID} and theorem \ref{ID-theorem}) as discussed in remark \ref{order-rem} is to work with \emph{unions} of consolidated districts directly instead of working with each single consolidated district at a time (and then having problems multiplying them in a ordered way). 
The corresponding ID-algorithm then iterates taking the ancestral closure and taking (the unions of) consolidated districts of the queried set until convergence. If the sets agree the causal effect is identifiable and the occuring products can be computed like in proposition \ref{id-eqn} point 3, with $D$ now a \emph{union} of consolidated districts.
The soundness then follows again with proposition \ref{id-eqn} and lemma \ref{new-central-lemma}, which also work in this case, but the algorithm might more often respond with ``FAIL''.
\end{Rem}

\end{document}